\documentclass[journal]{IEEEtran}

\usepackage{amsmath,amssymb,amsfonts}
\usepackage{algorithmic}
\usepackage{graphicx}
\usepackage{algorithm}
\usepackage{booktabs}
\usepackage{multirow}
\usepackage{tabularx}
\usepackage{ragged2e}
\usepackage{array}
\usepackage{makecell}
\usepackage{siunitx}
\usepackage{bm}

\usepackage[inline]{enumitem}
\usepackage[nocompress]{cite}
\usepackage{hyperref}
\hypersetup{
    colorlinks=true,
    linkcolor=blue,
    citecolor=blue,
    urlcolor=black,
    pdfborder={0 0 0}
}
\usepackage{hypcap}
\usepackage{textcomp}

\newcommand{\sig}{\rlap{\textsuperscript{*}}}

\def\BibTeX{{\rm B\kern-.05em{\sc i\kern-.025em b}\kern-.08em
    T\kern-.1667em\lower.7ex\hbox{E}\kern-.125emX}}

\begin{document}

\thispagestyle{plain} 
\pagestyle{plain}

\bstctlcite{IEEEtranBSTCTL}

\title{SingLEM: Single-Channel Large EEG Model}

\author{Jamiyan Sukhbaatar, 
        Satoshi Imamura, 
        Ibuki Inoue, 
        Shoya Murakami, 
        Kazi Mahmudul Hassan, 
        Seungwoo Han, 
        Ingon Chanpornpakdi, 
        and Toshihisa Tanaka

\thanks{This work was supported in part by JSPS KAKENHI 23H00548. The work of Jamiyan Sukhbaatar was supported by the Mongolia–Japan Engineering for Education Development (MJEED) project. (Corresponding author: Toshihisa Tanaka.)}
\thanks{Jamiyan Sukhbaatar is with the Department of Electronic and Information Engineering, Tokyo University of Agriculture and Technology, Koganei-Shi 184–8588, Japan, and also with the Department of Electronics and Communication Engineering, National University of Mongolia, Ulaanbaatar, 14200 Mongolia (e-mail: jamiyan@sip.tuat.ac.jp).}
\thanks{Satoshi Imamura, Ibuki Inoue, Shoya Murakami, Kazi Mahmudul Hassan, Seungwoo Han, Ingon Chanpornpakdi, and Toshihisa Tanaka are with the Department of Electronic and Information Engineering, Tokyo University of Agriculture and Technology, Koganei-Shi 184–8588, Japan. (e-mail: tanakat@cc.tuat.ac.jp)}}

\maketitle

\begin{abstract}
Current deep learning models for electroencephalography (EEG) are often task-specific and depend on large labeled datasets, limiting their adaptability. Although EEG foundation models seek broader applicability, many still rely on predefined multi-channel inputs, electrode-layout assumptions, or model-specific channel handling. To address these limitations, we introduce the Single-Channel Large EEG Model (SingLEM), a self-supervised foundation model whose hybrid convolutional--Transformer encoder maps each channel independently to a reusable representation capturing local and long-range temporal structure. These representations can be used individually or combined through channel-wise feature concatenation. We assembled 71 public EEG datasets comprising approximately 9,200 subjects and 357,000 single-channel hours. For leakage-controlled evaluation, downstream results were obtained with a model pretrained on 68 datasets after excluding the three source datasets underlying the six tasks. A model pretrained on all 71 datasets is provided for general reuse. Across six motor imagery and cognitive tasks under strict leave-one-subject-out (LOSO) evaluation, the leakage-controlled model with concatenated representations and a support vector machine (SVM) classifier achieved the best overall performance among the compared pretrained and classical feature-based methods. Additional classifier and subject-adapted analyses supported the robustness of its representations. These findings support single-channel self-supervised learning as a montage-flexible foundation for reusable EEG feature extraction and electrode-level spatial analysis. The source code and pretrained models are available at \url{https://github.com/ttlabtuat/SingLEM}.
\end{abstract}


\section{Introduction}
\label{sec:introduction}
\IEEEPARstart{E}{lectroencephalography} (EEG) is a non-invasive neurophysiological technique that measures brain activity through scalp electrodes. Because of its high temporal resolution, portability, and affordability, EEG is widely applied in diverse domains, including brain-computer interfaces (BCIs)~\cite{MI-BCI-popular-1}, sleep staging~\cite{Sleep-popular-1}, seizure detection~ \cite{Seizure-shoji2021automated, Seizure-2}, clinical diagnosis~\cite{Diagnosis-review-1}, and emotion recognition~\cite{Emotion-popular-1, Emotion-recent-1}. Despite its potential, EEG analysis is challenged by non-stationarity across subjects and sessions, susceptibility to noise (e.g., ocular or muscular artifacts), and low signal-to-noise ratios~\cite{EEG-challenges}. To address this, deep neural networks (DNNs) have emerged as the state-of-the-art paradigm, learning complex and task-relevant features automatically from raw data~\cite{DNN-for-EEG-review1}.

While DNNs have achieved remarkable success in various EEG decoding tasks, including motor imagery (MI) classification \cite{MI-BCI-recent-1, MI-BCI-recent-2}, steady-state visual evoked potentials (SSVEP) decoding \cite{SSVEP-BCI-1, SSVEPformer}, sleep stage classification \cite{Sleep-1, ito2025}, and cognitive workload estimation \cite{cognitive-2}, this task-specific end-to-end paradigm suffers from the following limitations.
First, these models are not generalizable and must be retrained for each new task. Second, they require substantial labeled data, which is costly and time-consuming to acquire. Third, the high inter-subject and inter-session variability inherent in EEG signals makes it difficult for these models to generalize to new users. These challenges highlight the urgent need for scalable, data-efficient, and generalizable EEG representation learning.

Inspired by the success of foundation models in natural language processing~\cite{devlin2019bert, achiam2023gpt} and computer vision~\cite{ViT}, recent studies have proposed foundation models for biomedical engineering and healthcare~\cite{FM_biomedical_eng, Foundation_model_review}. Such models leverage large-scale self-supervised pretraining to extract generalizable representations, which can be transferred to multiple downstream tasks with minimal fine-tuning. Their adaptability makes foundation models an attractive solution to the challenges of data scarcity and inherent signal variability in EEG analysis.

As a result, several EEG-specific foundation models have emerged, including BENDR \cite{BENDR}, BIOT \cite{BIOT}, LaBraM \cite{LaBraM}, CBraMod \cite{CBraMod}, CSBrain \cite{CSBrain}, LUNA \cite{LUNA}, CodeBrain \cite{CodeBrain}, and MIRepNet \cite{MIRepNet}. However, many multi-channel pretrained EEG models still require model-specific channel organization, electrode information, or adaptation procedures, which can complicate direct use as frozen feature extractors across heterogeneous or low-channel recordings. Although adaptation through fine-tuning is possible, it is a resource-intensive process that undermines the goal of having a universal, data-efficient model.

To overcome this challenge, instead of forcing a model to operate directly on multi-channel signals, we take a fundamentally different approach. EEG waveforms are montage dependent, and therefore, approaches that directly model multi-channel data introduce architectural rigidity and limit generalization. In contrast, our approach learns robust representations at the single-electrode level and concatenates the channel-wise representations to integrate them for downstream tasks. This design makes the model less dependent on fixed electrode layouts and well-suited for scenarios with heterogeneous or low-channel configurations.

Building on this motivation, we propose SingLEM, a self-supervised foundation model that learns EEG representations at the single-channel level. SingLEM is built on an asymmetric masked autoencoder (MAE) backbone~\cite{he2022masked} that integrates convolutional layers for local feature extraction with a hierarchical transformer to capture both short- and long-range temporal dependencies.

We assembled a large-scale corpus of 71 public EEG datasets comprising approximately 10,200 hours of multi-channel recordings, equivalent to 357,000 single-channel hours. We pretrained SingLEM on the complete corpus for general reuse. For the primary leakage-controlled evaluation, we pretrained a separate SingLEM model on 68 datasets after excluding the three source datasets used for the six downstream tasks. The resulting channel-level representations can be used individually or combined for downstream classification with lightweight classifiers, such as support vector machines (SVMs), without task-specific fine-tuning. Although spatial dependencies are not explicitly modeled during pretraining, spatial patterns can be recovered post hoc by aggregating per-channel representations.

Importantly, SingLEM is a single-channel representation learning framework, not a downstream classifier restricted to one electrode. The same pretrained encoder maps each EEG channel independently to a compact representation vector, and when multiple electrodes are available, these channel-wise representations can be combined through channel-wise feature concatenation. In the primary strict evaluation, we use simple feature concatenation followed by an SVM classifier to assess the frozen representations.

The main contributions of this study are summarized as follows:
\begin{itemize}
    \item We introduce SingLEM, a self-supervised foundation model that learns general-purpose representations from single-channel EEG, reducing dependence on fixed multi-channel electrode layouts.
    \item We show that channel-wise SingLEM representations can be combined through channel-wise feature concatenation for multi-channel downstream decoding, while preserving the flexibility of a single-channel encoder.
    \item In the primary strict LOSO evaluation, SingLEM (primary) achieves the best performance among the primary comparison methods across all six downstream EEG tasks when used as a frozen feature extractor with an SVM classifier.
    \item We establish SingLEM as a frozen feature extractor that avoids encoder fine-tuning while maintaining high downstream performance.
    \item We assemble 71 diverse datasets, totaling over 357,000 single-channel hours of recordings from approximately 9,200 subjects, and provide SingLEM (all 71 datasets) for general reuse.
\end{itemize}

\section{Related Work}
\subsection{Task-Specific Deep Learning for EEG Decoding}
Early deep learning approaches for EEG decoding mainly adopted end-to-end supervised architectures trained from scratch for individual tasks. Representative CNN-based models, such as DeepConvNet, ShallowConvNet~\cite{deep-convnet}, and EEGNet~\cite{EEGNet}, learn local temporal--spatial features directly from raw EEG using convolutional operations. Transformer-based models further extend this direction by using self-attention to capture longer-range temporal dependencies, as shown in EEG-Conformer~\cite{Conformer}, SSVEPformer~\cite{SSVEPformer}, and EEG-Deformer~\cite{Deformer}.

Despite their effectiveness, these supervised models are usually optimized for a particular task, dataset, and channel configuration. Consequently, they require substantial labeled data and are not readily reusable across different paradigms, subjects, or montages. This limitation has motivated a shift from task-specific supervised learning toward general-purpose pretrained EEG representation models.

\subsection{Self-Supervised Learning and Foundation Models for EEG}
To address task specificity and data scarcity, recent studies have explored self-supervised pretraining for EEG representation learning. The goal is to learn transferable EEG features from large unlabeled corpora and reuse them across downstream tasks with limited task-specific supervision.
 
Two major paradigms have emerged. Contrastive approaches, such as BENDR~\cite{BENDR} and BIOT~\cite{BIOT}, learn discriminative representations by bringing related EEG segments closer in the embedding space while separating unrelated segments. However, their performance depends strongly on how positive and negative pairs are constructed. Generative or masked-learning approaches, inspired by BERT~\cite{devlin2019bert} and MAE~\cite{he2022masked}, instead mask part of the signal or its tokenized representation and train the model to recover the missing content. LaBraM~\cite{LaBraM} and CBraMod~\cite{CBraMod} follow this direction using patch-based or masked-patch reconstruction strategies, while CodeBrain~\cite{CodeBrain}, CSBrain~\cite{CSBrain}, LUNA~\cite{LUNA}, and MIRepNet~\cite{MIRepNet} further explore structured, scalable, topology-aware, or MI-specific pretrained EEG representations.

Autoencoder-based representation learning has also been investigated for EEG, including AE-FBCSP~\cite{AEFBCSP} and GMAEEG~\cite{GMAEEG}. These studies support the usefulness of reconstruction-based objectives. In contrast, SingLEM learns frozen representations directly from individual EEG channels using a large heterogeneous corpus and combines available channels only at the downstream stage through channel-wise feature concatenation.

Recent systematic benchmarks have shown that EEG foundation models do not consistently outperform compact neural networks or classical non-neural decoders and that their relative performance depends strongly on the downstream task, adaptation strategy, and evaluation protocol~\cite{liu2026eeg, yang2026eeg}. Accordingly, comparing an EEG foundation model only with other pretrained models may not provide a complete assessment of its downstream performance relative to established classical and supervised neural decoders.

\subsection{Research Gap: Limited Adaptability of Existing Multi-Channel EEG Foundation Models}

The studies reviewed above provide a promising basis for general-purpose EEG representation learning. However, their practical use as fixed, ``ready-to-use'' feature extractors remains limited when downstream recordings differ in electrode layout, channel count, or available EEG channels. This limitation is especially important for low-channel, wearable, clinical, or heterogeneous EEG settings, where the available electrodes may not match the assumptions of a pretrained multi-channel model.

First, many existing pretrained EEG models remain coupled to channel layouts or electrode definitions used during pretraining. BENDR~\cite{BENDR} uses a fixed standard EEG channel configuration, and CBraMod~\cite{CBraMod} was pretrained using 19 common EEG channels. BIOT~\cite{BIOT} was pretrained with predefined bipolar and monopolar EEG channel configurations, whereas LaBraM~\cite{LaBraM} uses learnable electrode-specific positional embeddings. More recent models improve montage flexibility: LUNA, REVE, and DIVER-0 support variable electrode configurations through topology-agnostic channel unification, four-dimensional positional encoding, and channel-permutation-equivariant modeling, respectively~\cite{LUNA,REVE,DIVER0}, while CodeBrain, CSBrain, and MIRepNet provide scalable or task-specific pipelines~\cite{CodeBrain,CSBrain,MIRepNet}. Unlike these joint multi-channel approaches, SingLEM applies the same encoder independently to each channel.
 
Second, the generalizability of pretrained EEG models is constrained by the scope of their pretraining corpora. For instance, BIOT~\cite{BIOT} was pretrained on six public EEG datasets, while BENDR~\cite{BENDR} and CBraMod~\cite{CBraMod} primarily rely on the Temple University Hospital EEG Corpus. LaBraM~\cite{LaBraM} was pretrained on a smaller collection of around twenty public and self-collected EEG datasets. Such differences in corpus size, task diversity, recording conditions, and subject populations may affect transferability across recording setups, paradigms, and subject groups.

These limitations motivate an encoder independent of fixed electrode layouts. SingLEM addresses this gap by learning EEG representations at the single-channel level: the same pretrained encoder is applied independently to each available channel, while multi-channel information is combined only at the downstream stage through channel-wise feature concatenation. This design allows SingLEM to extract representations from the recorded channels without requiring a fixed encoder-level channel template or encoder retraining. Together, these properties provide a simple, montage-flexible foundation for EEG feature extraction across heterogeneous electrode configurations.

\begin{figure*}[!t]
    \centering 
    \includegraphics[width=\textwidth]{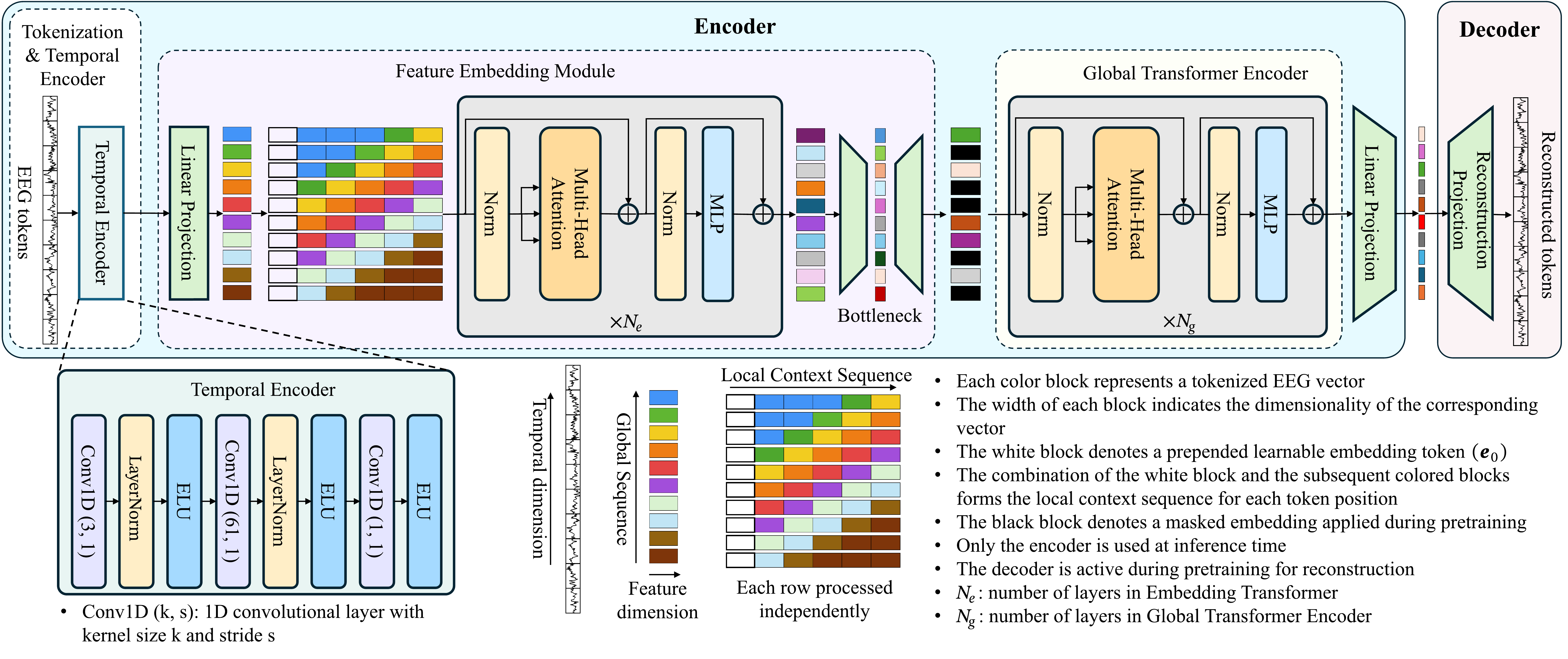} 
    \caption{Architecture of SingLEM. The model comprises three main modules: the temporal encoder, the feature embedding module, and the global transformer encoder. The temporal encoder extracts band-specific features from each EEG token independently. The feature embedding module enriches tokens with short-term contextual information using a sliding window mechanism. The global transformer encoder then integrates long-range dependencies across the sequence, yielding robust and context-aware representations.}
    \label{fig:model_architecture} 
\end{figure*}

\section{Method}
\label{sec:method}
This section outlines the proposed model architecture, data preprocessing, tokenization, formulation of the model, training procedure, and implementation details.

\subsection{Overview}
SingLEM is a self-supervised model for learning context-rich, robust representations from single-channel EEG signals without requiring fine-tuning for downstream tasks. SingLEM, as shown in Fig.~\ref{fig:model_architecture}, adopts an asymmetric MAE framework, employing a lightweight decoder and token masking applied solely during pretraining. The asymmetric design ensures that the encoder, rather than the decoder, is responsible for reconstructing the masked signals. By limiting the decoder's capacity, the model avoids the latent collapse and forces the encoder to capture informative and discriminative EEG features in its latent space. The encoder, trained to extract low-dimensional latent representations, is composed of three sequential modules: the temporal encoder, feature embedding module, and global transformer encoder. The lightweight CNN-based temporal encoder is introduced to capture frequency-specific features. The feature embedding module captures local temporal context, while the global transformer encoder integrates long-range dependencies across the EEG sequence.

\subsection{Data Collection and Preprocessing}
A large corpus of multi-channel EEG recordings was collected from the publicly available datasets listed in Table~\ref{tab:datasets}. Each dataset differs with respect to sampling rate, file format, montage, recording duration, subject cohort, and acquisition device. The corpus comprises 71 public EEG datasets from approximately 9,200 subjects, totaling 357,000 single-channel hours and spanning motor imagery, ERP/SSVEP, cognitive and affective paradigms, sleep, resting-state, and neurological EEG. Quality control included source-file accounting and readability checks and verification of subject and recording counts, channel metadata, sampling rates, and recording durations. All retained recordings underwent the common preprocessing described below. To unify datasets while preserving raw signal characteristics, we applied the following minimal preprocessing:
\begin{enumerate}
    \item Band-pass filtering (0.5--50 Hz): zero-phase FIR filtering to remove drift and high-frequency noise.
    \item Notch filtering: to attenuate powerline interference.
    \item Resampling to 128 Hz: to standardize temporal resolution across datasets.
    \item Artifact rejection: any sample that exceeds $\pm 100$ $\mu V$ is removed. Removed samples are treated as boundaries between continuous clean EEG segments.
    \item Scaling: all data are multiplied by $10^4$ so that amplitudes fall within the range (-1, 1).
\end{enumerate}

Following preprocessing, each channel is represented as one or more continuous clean EEG segments with uniform sampling and a bounded amplitude.

\subsection{EEG Tokenization}
To facilitate transformer-based processing, continuous clean EEG segments are divided into fixed-length overlapping tokens. This ensures a consistent input format for the transformer while preserving temporal context. Let a preprocessed clean EEG segment be $\mathbf{X}\,\in\mathbb{R}^{1\times S}$, where \(S\) is the total number of samples at 128 Hz. We define each token to cover \(\ell=128\) samples (1 s), with an overlap of \(u=32\) samples (250 ms)~\cite{DNN-for-EEG-review1} between successive tokens. Consequently, the stride is 
\(
s = \ell - u = 96
\). The number of tokens is 
\(
L = \big\lfloor (S - \ell)/s\big\rfloor + 1
\), 
and for \(i=0,\dots,L-1\), the $i$-th token $\mathbf{T}_i$ is extracted as
\begin{equation}
    \mathbf{T}_i \;=\;\mathbf{X}\bigl[i\,s: i\,s + \ell - 1\bigr]\;\in\;\mathbb{R}^{1\times \ell}.
\end{equation}

This overlapping scheme ensures each pair of adjacent tokens contains exactly \(u=32\) identical samples, thereby capturing transient events at the segment boundaries~\cite{DNN-for-EEG-review1}.

\subsection{Encoder Architecture}
The encoder is composed of three sequential modules, each designed to progressively capture and encode distinct levels of temporal abstraction from the input EEG signals. These three modules (the temporal encoder, feature embedding, and global transformer encoder modules) jointly learn hierarchical representations that reflect local spectral content, short-term contextual dependencies, and global sequence-level dynamics, respectively.

\paragraph*{Temporal Encoder}
Each EEG token \(\mathbf{T}_i\) is processed independently through a three-layer 1D CNN to extract band-specific temporal features, emulating band-pass filtering operations that span standard EEG frequency bands, from delta (0.5--4 Hz) to gamma (30--50 Hz). Each convolution layer is followed by normalization ($\mathrm{LN}$) and $\mathrm{ELU}$ activation to stabilize optimization and introduce nonlinearity. The first layer ($K^{(1)}{=}3$) captures high-frequency details, particularly in the gamma range. The second layer ($K^{(2)}{=}61$) targets slow oscillatory components, including delta activity. The final layer ($K^{(3)}{=}1$) acts as a channel-wise projection that consolidates features across filters while preserving the temporal length ($\ell$), producing per-token feature vectors for downstream embedding.

Let $\mathrm{Conv}^{(k)}$ denote the $k$‑th convolution layer, and define $\mathbf{U}_i^{(0)}=\mathbf{T}_i \in \mathbb{R}^{1 \times \ell}$. The processing can be expressed as
\begin{equation}
\mathbf{U}_i^{(k)} =
\begin{cases}
\mathrm{ELU}(\mathrm{LN}(\mathrm{Conv}^{(k)}(\mathbf{U}_i^{(k-1)}))),& k=1,2,\\
\mathrm{ELU}(\mathrm{Conv}^{(3)}(\mathbf{U}_i^{(2)})), & k=3,
\end{cases}
\end{equation}
where $\mathbf{U}_i^{(3)} \in \mathbb{R}^{1 \times \ell}$ serves as the initial token-level representation and is passed to the feature embedding module for contextualization.

\paragraph*{Feature Embedding Module} After token-level features have been extracted, a dedicated feature embedding module is applied to capture short-range temporal dependencies. This module produces contextualized embeddings for each token by integrating information from its temporal neighbors through a sliding window mechanism.

As the initial step, token-level features $\{\mathbf{U}_i^{(3)}\}$ are linearly projected from the CNN feature space into a learnable embedding space of dimension $d$ via $\mathbf{W}_E \in \mathbb{R}^{\ell \times d}$, which serves as an interface between the temporal encoder and subsequent transformer layers, expressed as
\begin{equation}
    \bm{\mathcal{V}}_i = \mathbf{U}_i^{(3)} \mathbf{W}_E.
\end{equation}
This projection yields a sequence of embedded tokens $\{\bm{\mathcal{V}}_i\}_{i=1}^{L}$, where each $\bm{\mathcal{V}}_i \in \mathbb{R}^{1 \times d}$. For each token position $i$, we construct a local context window of width $w$ centered on $\bm{\mathcal{V}_i}$ using boundary replication for edge padding, as follows
\begin{equation}
\mathbf{S}_i = \left[\bm{\mathcal{V}}_{i-\lfloor w/2 \rfloor}, \dots, \bm{\mathcal{V}}_{i+\lfloor w/2 \rfloor}\right] \in \mathbb{R}^{w \times d}.
\end{equation}
For each local context window $\mathbf{S}_i$, we prepend a learnable embedding token \(\mathbf{e}_0\in\mathbb{R}^{d}\) to form the local context sequence, as follows:
\begin{equation}
    \widetilde{\mathbf{S}}_i = [\mathbf{e}_0; \mathbf{S}_i] \in \mathbb{R}^{(w+1) \times d}.
\end{equation}
To preserve temporal order, a positional embedding matrix $\mathbf{P}\in\mathbb{R}^{(w+1)\times d}$ is added element-wise, as follows:
\begin{equation}
    \mathbf{H}_i = \widetilde{\mathbf{S}}_i +\mathbf{P}.
\end{equation}

Each local context sequence $\mathbf{H}_i$ is independently processed by a lightweight transformer with $N_e$ layers. Unlike typical global transformers, our design ensures that each $\mathbf{H}_i$ models only local context. We implement this by reshaping the input batch as $(\texttt{batch\_size} \times L)$, allowing all $L$ local windows to be processed in parallel. This design prioritizes the efficient modeling of local dependencies before global integration.

Each transformer layer contains a multi-head self-attention (MHSA) block and a feed-forward network (FFN), both wrapped with residual connections and layer normalization as follows
\begin{align}
\mathbf{A}_i &= \mathrm{MHSA}(\mathrm{LN}(\mathbf{H}_i)) + \mathbf{H}_i, \\
\mathbf{Z}_i &= \mathrm{FFN}(\mathrm{LN}(\mathbf{A}_i)) + \mathbf{A}_i.
\end{align}
The FFN comprises two linear layers with an intermediate GELU activation, projecting the dimension as $d \rightarrow 4d \rightarrow d$.

After $N_e$ layers, the output corresponding to the prepended token (index 0) is selected as the contextual embedding $\mathbf{z}_i \in \mathbb{R}^{d}$. This embedding is then passed through an MLP-based bottleneck as
\begin{equation}
\begin{split}
\mathbf{e}_i &= \mathbf{W}_{b3}\,\mathrm{GELU}\left(\mathbf{W}_{b2}\,\mathrm{GELU}\left(\mathbf{z}_i \mathbf{W}_{b1} + \mathbf{b}_1\right) + \mathbf{b}_2\right) \\
&\quad + \mathbf{b}_3,
\end{split}
\end{equation}
where $\mathbf{W}_{b1} \in \mathbb{R}^{d \times \tfrac{d}{2}}, \mathbf{W}_{b2} \in \mathbb{R}^ \mathrm{\tfrac{d}{2} \times {d_{emb}}}, \mathbf{W}_{b3} \in \mathbb{R}^\mathrm{{d_{emb}} \times D}$, and $d_\mathrm{emb} < d$. This bottleneck encourages the representation to retain only the most salient features from each local context window, producing compact and highly informative embeddings.

\paragraph*{Global Transformer Encoder} To model long-range temporal dependencies across the entire EEG sequence, the contextualized embeddings $\{\mathbf{e}_i\}_{i=1}^L$ are stacked into $\mathbf{E} = [\mathbf{e}_1;\dots;\mathbf{e}_L]\in\mathbb{R}^{L\times D}$, and augmented with learnable positional embeddings $\mathbf{Q}\in\mathbb{R}^{L\times D}$ as follows:
\begin{equation}
    \mathbf{G}^{(0)} = \mathbf{E} + \mathbf{Q}.
    \label{eq: EMBs}
\end{equation}
This sequence $\mathbf{G}^{(0)}$ is processed by a transformer encoder with $N_g$ layers, each comprising an MHSA block followed by an FFN, both with pre-layer normalization and residual connections. Unlike the feature embedding module, no MLP bottleneck is applied, and the attention is computed globally across all tokens.

After the final layer, the outputs $\mathbf{G}^{(N_g)} \in \mathbb{R}^{L \times D}$ are projected via $\mathbf{W}_R \in \mathbb{R}^{D \times r}$ to produce the following final per-token representations:
\begin{equation}
\mathbf{R} = \mathbf{G}^{(N_g)} \mathbf{W}_R.
\end{equation}
These per-token representations $\mathbf{R} \in \mathbb{R}^{L \times r}$ serve as the latent features used for downstream tasks such as classification or regression.

This hierarchical design---processing from local temporal filtering, to short-term contextualization, and finally to global integration---is designed to capture multi-scale dependencies within the EEG token sequences.

\subsection{Pretraining SingLEM}
We employ a self-supervised MAE paradigm when pretraining SingLEM.
Each pretraining batch comprises randomly sampled sequences of consecutive EEG tokens extracted from the preprocessed data. Each sequence is sampled within a single continuous clean segment, ensuring that it remains within the same channel and recording. Although sequence lengths can vary across batches, each batch has a uniform length for computational consistency.

The pretraining pipeline consists of the following stages:

\paragraph*{Masking} From the feature embedding module, we obtain per-token contextual embeddings $\mathbf{E} = [\mathbf{e}_1;\dots;\mathbf{e}_L] \in \mathbb{R}^{L\times D}$. A fixed proportion $p = 0.5$ of these token embeddings is randomly selected for masking. Let $\mathcal{M} \subset \{1, \dots, L\}$ denote the set of masked indices. The masked sequence is then
\begin{equation}
\widetilde{\mathbf{e}}_i =
\begin{cases}
\mathbf{0}, & i \in \mathcal{M} \\
\mathbf{e}_i, & \text{otherwise},
\end{cases}
\end{equation}
yielding $\widetilde{\mathbf{E}}=[\widetilde{\mathbf{e}}_1,\dots,\widetilde{\mathbf{e}}_L] \in \mathbb{R}^{L \times D}$.

\paragraph*{Global transformer encoding and projection} To encode sequence order, learnable positional embeddings $\mathbf{Q} \in \mathbb{R}^{L \times D}$ are added using $\mathbf{G}^{(0)} = \widetilde{\mathbf{E}} + \mathbf{Q}$. The result is then processed by the global transformer encoder, and its final outputs $\mathbf{G}^{(N_g)}$ are linearly projected to the encoder's final representations $\mathbf{R} = [\mathbf{r}_1; \dots; \mathbf{r_L}] \in \mathbb{R}^{L \times r}$.

\paragraph*{Reconstruction} Each representation $\mathbf{r}_i$ is mapped back to the original token space using a shared lightweight decoder as follows:
\begin{equation}
    \widehat{\mathbf{T}}_i = \mathbf{r}_i \mathbf{W}_{\mathrm{dec}} + \mathbf{b}_{\mathrm{dec}}, \quad \mathbf{W}_{\mathrm{dec}} \in \mathbb{R}^{r \times \ell}, \;\; \widehat{\mathbf{T}}_i \in \mathbb{R}^{1 \times \ell}.
\end{equation}
Unlike conventional MAE approaches \cite{devlin2019bert, he2022masked}, which reconstruct only the masked tokens, we jointly minimize the reconstruction loss over both masked and unmasked tokens. This ensures that the encoder’s compressed representations preserve all the information for downstream use.

\paragraph*{Loss computation}
We optimize the following three loss terms:
\begin{equation}
    \mathcal{L}_{\mathrm{masked}} = \frac{1}{|\mathcal{M}|} \sum_{i\in\mathcal{M}} \mathrm{Huber}\left(\mathbf{T}_i,\widehat{\mathbf{T}}_i\right),
\end{equation}

\begin{equation}
    \mathcal{L}_{\mathrm{unmasked}} = \frac{1}{L-|\mathcal{M}|}\sum_{i\notin\mathcal{M}}
    \mathrm{Huber}\left(\mathbf{T}_i,\widehat{\mathbf{T}}_i\right),
\end{equation}

\begin{equation}
    \mathcal{L}_{\beta\gamma} = \frac{1}{L}\sum_{i=1}^L
    \left\lVert \mathrm{BP}_{13-50}(\mathbf{T}_i) - \mathrm{BP}_{13-50}(\widehat{\mathbf{T}}_i)\right\rVert_2^2,
\end{equation}
where $\mathrm{Huber}(\cdot)$ denotes the Huber loss function and $\mathrm{BP}_{13-50}(\cdot)$ denotes a zero-phase filter retaining frequencies in the 13--50 Hz range. The total pretraining loss is
\begin{equation}
    \mathcal{L} = \lambda_1\,\mathcal{L}_{\mathrm{masked}}
 + \lambda_2\,\mathcal{L}_{\mathrm{unmasked}}
 + \lambda_3\,\mathcal{L}_{\beta\gamma},
\end{equation}
with $\lambda_1,\lambda_2,$ and $\lambda_3$ controlling the contribution of each term.

After convergence, the decoder and mask module are discarded, and the final encoder can now be used as a feature extractor for downstream tasks.

\begin{table*}[t!]
\centering
\caption{Summary of the complete 71-dataset pretraining corpus.}
\label{tab:datasets}
\footnotesize
\begin{tabular}{@{} l l p{4cm} r r r S[table-format=4.2] l @{}}
\toprule
\textbf{Dataset (Year)} & \textbf{Ref.} & \textbf{Task / Type} & \textbf{Subj.} & \textbf{Ch.} & \textbf{Rate [Hz]} & \textbf{Duration [h]} & \textbf{Format} \\
\midrule
Cho (2017) & \cite{cho-h} & MI & 52 & 64 & 512 & 26.20 & MAT \\
Schalk (2009) & \cite{schalk-g} & MI & 109 & 64 & 160 & 49.00 & EDF \\
Shin (2017) & \cite{shin-j-3} & MI & 29 & 30 & 200 & 29.00 & MAT \\
Kaya (2018) & \cite{kaya-m} & MI & 13 & 22 & 1000 & 67.60 & MAT \\
Lee (2019) & \cite{lee-m} & MI+ERP+SSVEP & 54 & 62 & 1000 & 255.10 & MAT \\
Brandl (2020) & \cite{brandl-s} & Cognitive: Stroop & 16 & 63 & 1000 & 19.75 & MAT \\
Chen (2023) & \cite{chen-z} & Visual: Color-Word & 21 & 34 & 1000 & 1.32 & Neuroscan \\
Mou (2024) & \cite{mou-x} & Cognitive & 10 & 64 & 1000 & 166.16 & EDF \\
Getzmann (2024) & \cite{getzmann-s} & Cognitive & 608 & 128 & 1000 & 118.85 & BrainVision \\
Ji (2024) & \cite{ji-x} & Cognitive & 80 & 64 & 1000 & 186.16 & MAT \\
Momenian (2024) & \cite{momenian-m} & Auditory: Story & 52 & 64 & 1000 & 18.95 & EEGLAB \\
Xiang (2024) & \cite{xiang-c} & Sleep & 71 & 61 & 500 & 17.75 & EEGLAB \\
Babayan (2019) & \cite{babayan-a} & Sleep & 227 & 62 & 500 & 60.17 & BrainVision \\
van Dijk (2022) & \cite{van-dijk} & ERP & 1,274 & 22 & 2500 & 88.86 & BrainVision \\
Ngo (2024) & \cite{ngo-t} & Cognitive: Spelling & 176 & 34 & 1000 & 66.32 & EDF \\
Grootswagers (2022) & \cite{grootswagers-t-16} & Visual: RSVP & 50 & 128 & 1000 & 41.73 & BrainVision \\
Dzianok (2024) & \cite{dzianok-p} & Memory & 79 & 128 & 1000 & 44.51 & BrainVision \\
Ma (2022) & \cite{ma-j} & MI & 25 & 32 & 250 & 12.99 & MAT \\
Chen (2024) & \cite{chen-y} & Cognitive: Deception & 24 & 31 & 500 & 64.72 & BrainVision \\
Cao (2019) & \cite{cao-z} & Cognitive: Sustained & 27 & 32 & 500 & 81.94 & EEGLAB \\
Telesford (2023) & \cite{telesford-q} & Visual: Movie & 22 & 64 & 5000 & 5.90 & EEGLAB \\
Ma (2020) & \cite{ma-x} & MI & 25 & 64 & 500 & 45.41 & Neuroscan \\
Yang (2025) & \cite{yang-b} & MI & 62 & 64 & 1000 & 127.83 & Biosemi \\
Chen (2022) & \cite{chen-k} & Cognitive & 31 & 128 & 1000 & 40.10 & EEGLAB \\
Dreyer (2023) & \cite{dreyer-p} & MI & 87 & 27 & 512 & 71.78 & GDF \\
Wang (2022) & \cite{wang-y} & Memory & 60 & 63/64 & 500 & 74.98 & BrainVision \\
Chen (2023) & \cite{chen-j} & Emotion Recognition & 123 & 32 & 1000 & 155.28 & Biosemi \\
Shin (2018) & \cite{shin-j-28} & Cognitive & 26 & 28 & 1000 & 40.19 & BrainVision \\
Hinss (2023) & \cite{hinss-m} & Cognitive: Vigilance & 29 & 64 & 500 & 81.04 & EEGLAB \\
Gebodh (2021) & \cite{gebodh-n} & Cognitive: Vigilance & 26 & 32 & 1000 & 72.78 & EEGLAB \\
Nieto (2022) & \cite{nieto-n} & Inner Speech & 10 & 128 & 1024 & 12.95 & Biosemi \\
Won (2022) & \cite{won-k} & P300 & 55 & 32 & 512 & 44.09 & EEGLAB \\
Hollenstein (2022) & \cite{hollenstein-n} & Reading & 30 & 128 & 500 & 43.13 & MAT \\
Liu (2025) & \cite{liu-y} & MI & 27 & 64 & 1000 & 33.45 & EEGLAB \\
Wagner (2019) & \cite{wagner-j} & Motor: Walking & 20 & 108 & 512 & 39.84 & EEGLAB \\
Ghosh (2022) & \cite{ghosh-r} & Cognitive: Stroop & 40 & 32 & 128 & 3.30 & MAT \\
Valdes-Sosa (2021) & \cite{valdes-sosa} & Clinical Activation Protocol & 282 & 64/120 & 200 & 107.90 & EDF \\
Mheich (2021) & \cite{mheich-a} & Cognitive & 43 & 256 & 1000 & 24.74 & BrainVision \\
Liu (2024) & \cite{liu-h} & MI & 50 & 29 & 500 & 4.44 & EDF \\
Choi (2024)& \cite{choi-g} & Emotion (Voice) & 44 & 63 & 1000 & 46.55 & MAT \\
Wei (2024) & \cite{wei-x} & Sleep & 29 & 19 & 1000 & 217.78 & EDF \\
Gu (2024) & \cite{gu-m} & SSVEP & 30 & 64 & 1000 & 235.78 & EDF \\
Mumtaz (2017) & \cite{mumtaz-w} & ERP: Oddball & 64 & 19 & 256 & 4.44 & MAT \\
Iwama (2023) & \cite{iwama-s} & Sensorimotor Rhythm & 138 & 128 & 1000 & 154.28 & EDF \\
Liwicki (2023) & \cite{simistira-l} & Inner Speech & 4 & 64 & 512 & 1.47 & Biosemi \\
Moffa (2022) & \cite{moffa-a} & Brain Stimulation & 24 & 64 & 2048 & 125.58 & MAT \\
Cuevas-Romero (2022) & \cite{cuevas-r} & Auditory: Tinnitus & 89 & 17 & 256 & 99.17 & EEGLAB \\
Pei (2022) & \cite{pei-x} & Cognitive: Alertness & 74 & 64 & 1000 & 110.16 & Curry \\
Lin (2025) & \cite{lin-n} & Epilepsy (IED) & 52 & 19 & 500 & 28.00 & MAT \\
Pavlov (2022) & \cite{pavlov-y} & Cognitive: Memory & 65 & 64 & 1000 & 140.73 & EEGLAB \\
Liu (2022) & \cite{liu-b} & SSVEP & 100 & 64 & 1000 & 19.68 & EDF \\
Cai (2022) & \cite{cai-h} & Visual: Faces & 53 & 128/3 & 250 & 16.57 & MAT/RAW \\
Pascucci (2022) & \cite{pascucci-d} & Visual: Face ID & 20 & 128 & 2048 & 10.57 & Biosemi \\
Stieger (2021) & \cite{stieger-j} & MI & 62 & 64 & 1000 & 613.34 & MAT \\
Lopez (2017) & \cite{lopez-s} & Abnormal & 2,383 & Varies & 250 & {1,141.00} & EDF \\
Buckwalter (2021) & \cite{buckwalter-g} & Artifact & 213 & Varies & 250 & 99.98 & EDF \\
Veloso (2017) & \cite{veloso-l} & Epilepsy & 200 & Varies & 250 & 631.84 & EDF \\
Harati (2015) & \cite{harati-a} & Event & 370 & Varies & 250 & 129.49 & EDF \\
Shah (2018) & \cite{shah-v} & Seizure & 675 & Varies & 250 & {1,473.57} & EDF \\
von Weltin (2017) & \cite{von-w} & Slowing & 38 & Varies & 250 & 27.60 & EDF \\
Grootswagers (2025) & \cite{grootswagers-t-61} & Visual Perception & 42 & 32 & 500 & 5.52 & EEGLAB \\
Yi (2025) & \cite{yi-w} & MI & 18 & 64 & 1000 & 32.48 & Neuroscan \\
Rybář (2025) & \cite{rybavr-m} & Mental Task & 19 & 64 & 2048 & 22.29 & Biosemi \\
He (2025) & \cite{he-b} & Visual: Object ID & 40 & 32 & 500 & 23.73 & EDF \\
Xue (2025) & \cite{xue-s} & Visual: RSVP & 32 & 64 & 1000 & 157.93 & Curry \\
Bai (2025) & \cite{bai-y} & Reading & 30 & 64 & 1000 & 12.92 & Biosemi \\
Moreira (2025) & \cite{moreira-j} & Speech Decoding & 24 & 64 & 1000 & 50.76 & EDF \\
Wang (2025) & \cite{wang-q} & Auditory: Listening & 26 & 64 & 500 & 19.20 & BrainVision \\
Zhang (2025) & \cite{zhang-g} & Object Viewing & 30 & 64 & 500 & 23.70 & EEGLAB \\
Tao (2024) & \cite{tao-x} & Driving Task & 35 & 59 & 1000 & 55.07 & Biosemi \\
López-Larraz (2025) & \cite{Lopez-Larraz} & Sleep & 128 & 2/6 & 256 & {2,002.59} & EDF \\ 
\bottomrule
\end{tabular}
\end{table*}

\subsection{Implementation Details}

The CNN-based temporal encoder consisted of three convolutional layers, $\mathrm{Conv}^{(1)}$, $\mathrm{Conv}^{(2)}$, and $\mathrm{Conv}^{(3)}$, with output channels set to 32, 32, and 1, respectively. The feature vectors from the temporal encoder were individually projected into a $d = 128$ dimensional space. For local contextualization, a temporal window of $w = 5$ within the contextual embedding module was used.
The feature embedding transformer consisted of $N_e = 4$ layers with four attention heads each. The subsequent MLP bottleneck projected the token embeddings to $d_{\mathrm{emb}} = 32$. The global transformer encoder was implemented with $D = 128$, $N_g = 12$ layers, and eight attention heads per layer. A final linear projection of each token resulted in an output with a compact representation of size $r = 16$.

SingLEM was implemented in PyTorch and trained for 16 epochs on four NVIDIA A100 GPUs using distributed data parallelism. The optimizer was AdamW ($\beta_1 = 0.9$, $\beta_2 = 0.95$, and a weight decay of $10^{-1}$) with a cosine annealing learning rate schedule that decayed from $10^{-4}$ to $10^{-6}$, and a batch size of 1024.

\subsection{SingLEM Configurations}

We trained three SingLEM configurations. \emph{SingLEM (primary)} uses the complete architecture and was pretrained on 68 datasets after excluding the three downstream source datasets. \emph{SingLEM (all 71 datasets)} uses the same architecture and training procedure on the complete corpus and is provided for general reuse. \emph{SingLEM (w/o feature emb.)} was pretrained on the same 68 datasets as SingLEM (primary) but omits the feature embedding module. Unless otherwise stated, ``SingLEM'' refers to SingLEM (primary); the other two configurations are reported as ablation variants.

\section{Experiments}
\label{sec:experiments}
This section outlines the experimental framework used to evaluate SingLEM. We describe the compared pretrained models, classical EEG feature baselines, supervised neural baselines, datasets, preprocessing procedures, strict and subject-adapted evaluation protocols, implementation details, and performance metrics.

\subsection{Baselines}
We benchmarked SingLEM against three groups of baselines: pretrained EEG models, classical EEG feature baselines, and supervised neural decoders.

\subsubsection{Foundation Models}
We compared SingLEM with eight pretrained EEG models: BENDR~\cite{BENDR}, BIOT~\cite{BIOT}, LaBraM~\cite{LaBraM}, CBraMod~\cite{CBraMod}, CodeBrain~\cite{CodeBrain}, CSBrain~\cite{CSBrain}, LUNA-large~\cite{LUNA}, and MIRepNet~\cite{MIRepNet}. For the primary strict LOSO evaluation, SingLEM and all baseline pretrained EEG models were kept frozen and used only as feature extractors.
   
Because the pretrained models require different input-channel handling procedures during feature extraction, each baseline was evaluated using the input format and channel-handling procedure required by its pretrained implementation. When expected electrodes were unavailable, we constructed the model input using predefined nearest-channel substitution, retaining available matched channels, zero-filled channel entries, or interpolation, as required by the model. MIRepNet was evaluated only on the MI datasets because it is designed for MI decoding and uses an MI-specific preprocessing and channel-alignment pipeline.

\subsubsection{Classical EEG Features}
As traditional EEG baselines, we evaluated common spatial pattern (CSP) and Welch power spectral density (PSD) features. CSP was included as a classical spatial-filtering baseline. Although it is  particularly well established for MI decoding, it was evaluated on all six downstream tasks under the same strict LOSO protocol. In each LOSO fold, CSP was fitted only on the corresponding training partition after 8--30 Hz band-pass filtering, and up to six CSP components were used to extract log-variance features.

Welch PSD was included as a stable frequency-domain baseline. For each trial, log-PSD features were computed independently for each channel in the 4--40 Hz range and flattened into a channel-frequency feature vector. For both CSP and Welch PSD, feature scaling was fitted only on the corresponding training partition before downstream classification.

\subsubsection{Supervised Neural Decoders}
To compare SingLEM with supervised neural EEG decoders, we additionally evaluated EEGNet~\cite{EEGNet}, EEGConformer~\cite{Conformer}, and IFNetV2~\cite{IFNet}. In the strict LOSO evaluation, these models were trained directly on the downstream labeled data using the source-subject training and validation splits defined for each LOSO fold.

\subsection{Datasets and Preprocessing}

\subsubsection{Datasets}
We evaluated SingLEM on six diverse EEG classification tasks: three MI and three cognitive tasks, derived from three publicly available EEG datasets.

\paragraph*{MI Datasets}

\begin{enumerate*}[label=\emph{\alph*})]
\item \textbf{Dreyer-MI-2C}~\cite{dreyer-p}: This dataset contains two-class MI (left vs. right hand) recordings. The original publication provides three subsets (A, B, and C). To reduce computational overhead, we utilized Dataset B for this study, which contains data from 21 participants. EEG data were recorded using a g.USBamp (g.tec, Austria) amplifier from 27 active electrodes placed according to the international 10--20 system and a sampling rate of 512 Hz. Each subject completed six runs, of which the last four runs were designated for user training. Each trial lasted 8 s, initiated by a visual cue at $t=3$ s. Following standard BCI protocol, we extracted the 5-s segment from $t=3$ s to $t=8$ s corresponding to the active MI period. We utilized data from the four user-training runs, yielding 160 trials per subject.
\item \textbf{WBCIC-MI-2C}~\cite{yang-b}: This large-scale two-class MI dataset (left- vs. right-hand grasping) was collected from 51 healthy subjects during the 2019 World Robot Conference Contest-BCI Robot Contest MI (WBCIC-MI). Data were recorded with a 64-channel wireless EEG system (Neuracle, China) using 59 EEG channels (10--20 montage) at 1,000 Hz. Each 7.5-second trial included a 4-s MI period. We used all 200 trials from the first recording session for each subject.
\item \textbf{WBCIC-MI-3C}~\cite{yang-b}: This dataset is also from the WBCIC-MI collection, and it contains three-class MI (left-hand grasping, right-hand grasping, and foot hooking) recordings from 11 healthy subjects. The recording setup and trial structure were identical to those of the WBCIC-MI-2C. We used all 300 trials from the first session for each subject.
\end{enumerate*}

\paragraph*{Cognitive Task Datasets}

The following three datasets were obtained from a study on simultaneous EEG-NIRS acquisition during cognitive tasks~\cite{shin-j-28}. Data for all three tasks were acquired from 26 healthy participants using a BrainAmp amplifier (Brain Products GmbH, Germany). EEG was recorded from 30 electrodes (10–5 montage), with two used for reference and ground, yielding 28 channels for analysis. For all tasks, we extracted 10-s trials.
\begin{enumerate*}[label=\emph{\alph*})]
\item \textbf{N-back-2C}: This cognitive dataset was recorded during an n-back task designed to assess working memory. The original protocol involved 40-s task blocks and 20-s rest blocks. To create balanced trials, the initial 20 s of each task block and the full 20-s rest blocks were used. These 20-s segments were split into 10-s trials, resulting in 108 trials per subject. The classification task for this dataset was to distinguish between the working memory task and the rest condition.

\item \textbf{DSR-2C}: The Discrimination/Selection Response (DSR) dataset was collected during tasks requiring stimulus discrimination followed by response selection, which is a measure of cognitive attention. The protocol was similar to that of the N-back dataset, with 40-s task blocks and 20-s rest periods. The initial 20 s of each task block were segmented into two 10-s trials, along with the rest of the blocks, yielding a total of 72 trials per subject. Cognitive attention and rest classification were performed.

\item \textbf{WG-2C}: In this dataset, participants performed word-generation (WG) and baseline (BL) tasks. During WG, subjects silently generated words starting with a given letter for 10 s, avoiding repetitions. The BL task involved relaxing and gazing at a fixation cross for 10 s. This study focused on classifying high versus low cognitive load corresponding to WG and BL, with 60 trials per participant.
\end{enumerate*}

\subsubsection{Preprocessing}
To ensure a fair comparison, EEG preprocessing was matched to the input requirements specified by each pretrained model or official implementation. Continuous EEG recordings were notch-filtered at 50 Hz, band-pass filtered, and resampled before trial extraction. Trials prepared for SingLEM, Welch PSD, and the supervised neural decoders used 0.5--50 Hz filtering and 128 Hz resampling; CSP features were computed from 8--30 Hz band-pass-filtered trials.
   
For the pretrained baseline models, model-specific preprocessing settings were used. BENDR used 0.5--50 Hz filtering at 256 Hz; BIOT used 0.5--50 Hz filtering at 200 Hz; LaBraM and CSBrain used 0.1--75 Hz filtering at 200 Hz; CBraMod and CodeBrain used 0.3--75 Hz filtering at 200 Hz; LUNA used 0.1--75 Hz filtering at 256 Hz; and MIRepNet used 8--30 Hz filtering at 250 Hz and was applied only to the MI datasets.
 
After trial extraction, signals were converted to the input scale required by each model. SingLEM, LaBraM, CBraMod, CodeBrain, and CSBrain used trials in microvolt units scaled by $10^{-2}$; LUNA used per-trial per-channel z-score normalization; BENDR used relative-amplitude scaling during feature extraction; BIOT used percentile-based channel normalization; and MIRepNet used its MI-specific spatial alignment procedure. Any feature scaling used by downstream classifiers was fitted only on the corresponding training partition.

\subsection{Evaluation Protocol}

We used two LOSO-based evaluation protocols. In the primary strict LOSO protocol, one subject was held out as the test subject, while data from the remaining subjects were used as source-subject data. The source-subject data were split into stratified 80\% training and 20\% validation sets for model selection and hyperparameter tuning. After model selection, the final classifier or neural decoder was trained using all source-subject data and evaluated on the held-out subject. Target-subject trials were not used for feature scaling, hyperparameter selection, epoch selection, or training. Because SingLEM (primary) excluded the three downstream source datasets during pretraining, no subject, recording, or trial from these datasets was observed during pretraining.

As an additional analysis, we also performed a partially subject-adapted LOSO evaluation. For each held-out subject, 30\% of the trials in each class were randomly selected using a subject-level random-number generator initialized with seed 2023, yielding a class-balanced calibration set; all remaining trials were reserved exclusively for final testing. Hyperparameters and epoch selection were still determined using only source-subject training and validation data. For feature-based classifiers, the MLP classifier was trained using the source-subject data together with the target-subject calibration trials. For supervised neural decoders, the source-trained backbone was frozen, and only the existing classifier head was adapted using the target calibration trials. Target-test trials were not used for feature scaling, hyperparameter selection, epoch selection, training, or adaptation, and were reserved only for final evaluation.

\subsection{Implementation Details}
For feature-based evaluation, we used RBF-SVM classifiers implemented with RAPIDS cuML and MLP classifiers implemented with PyTorch. SVM hyperparameters were selected on the source-validation set using macro-F1 over 40 Optuna trials, with $C$ selected from 20 logarithmically spaced values between $10^{-1}$ and $10^{2}$ and $\gamma \in \{\texttt{scale},\texttt{auto}\}$. The MLP comprised one 64-unit hidden layer with ReLU activation and dropout ($p=0.1$), and was trained with AdamW, a learning rate of $10^{-4}$, weight decay of $10^{-5}$, batch size of 128, cosine scheduling, and early stopping. All models followed the data-use restrictions defined in the Evaluation Protocol. Full hyperparameters and implementation details are provided in the project repository.

\subsection{Evaluation Metrics}
The primary strict LOSO SVM evaluation reports accuracy, macro-F1, and Cohen's $\kappa$, whereas the MLP, supervised neural decoder, and subject-adapted analyses are summarized by accuracy. Accuracy was the primary inferential metric, whereas macro-F1 and Cohen's $\kappa$ were retained as complementary descriptive metrics. Within each dataset and evaluation setting, two-sided paired Wilcoxon signed-rank tests compared SingLEM (primary) with each applicable non-SingLEM baseline using held-out-subject accuracies; Holm correction was applied across these baseline comparisons, with statistical significance defined as $p_{\mathrm{Holm}}<0.05$.

\section{Results}
\label{sec:results}
 
This section first reports the primary strict LOSO SVM evaluation using frozen pretrained feature extractors and classical EEG features. We then present MLP, supervised neural decoder, and partially subject-adapted analyses, followed by single-channel spatial analysis. Finally, we examine the effects of downstream-dataset exposure during pretraining and removal of the feature embedding module; these ablation variants are also reported in Tables~\ref{tab:MI_results} and~\ref{tab:Cognitive_results}.

\subsection{Primary Strict LOSO Results with SVM}
 
We evaluated SingLEM using multi-channel EEG information by concatenating independently extracted channel-wise features. The resulting representations were compared with baseline features across three MI and three cognitive tasks. In this primary comparison, pretrained EEG feature extractors were kept frozen, and all feature representations, including classical EEG features, were evaluated using the same SVM classifier protocol. The SingLEM ablation variants are included at the bottom of Tables~\ref{tab:MI_results} and~\ref{tab:Cognitive_results} but are not treated as baseline methods in the primary performance ranking.

\subsubsection{Performance on MI Tasks}
 
The results for the three MI tasks are summarized in Table~\ref{tab:MI_results}. The primary SingLEM model achieved the best performance among the compared baseline methods across all three MI datasets. On the \textbf{Dreyer-MI-2C} dataset, SingLEM achieved a mean accuracy of 74.58\% and a macro-F1 of 74.45\%, outperforming the strongest baseline, MIRepNet, by approximately 1.9 percentage points in accuracy. On the three-class \textbf{WBCIC-MI-3C} task, SingLEM achieved an accuracy of 68.14\% and a macro-F1 of 68.03\%, exceeding the next-best baseline, CSBrain, by more than 5 percentage points in accuracy. On the \textbf{WBCIC-MI-2C} dataset, SingLEM achieved 79.68\% accuracy and a 79.60\% macro-F1, slightly outperforming CSBrain. Across all three MI tasks, the primary SingLEM model also obtained the highest Cohen's $\kappa$ values among the compared baseline methods, indicating stronger agreement beyond chance.

\subsubsection{Performance on Cognitive Tasks}
 
As shown in Table~\ref{tab:Cognitive_results}, the primary SingLEM model also achieved the best performance among the compared baseline methods across all three cognitive tasks in the primary strict LOSO comparison. On the \textbf{N-back-2C} task, SingLEM achieved 84.15\% accuracy and an 83.57\% macro-F1. On the \textbf{DSR-2C} task, SingLEM achieved 85.68\% accuracy and an 85.46\% macro-F1, with a Cohen's $\kappa$ of 0.714. On the \textbf{WG-2C} task, SingLEM achieved the highest accuracy of 70.26\%, although the margin over CBraMod was small. These results show that SingLEM provides transferable representations across both MI and cognitive EEG classification tasks.

\begin{table*}[!t]
\centering
\caption{Strict LOSO results on three MI datasets (mean $\pm$ sd across held-out subjects). Best means among the primary comparison methods are bold; ablation variants are excluded from the ranking.}
\label{tab:MI_results}
\scriptsize
\setlength{\tabcolsep}{3pt}
\begin{tabular}{@{}l ccc ccc ccc@{}}
\toprule
 
\multirow{2}{*}{\textbf{Model}} & \multicolumn{3}{c}{\textbf{Dreyer-MI-2C}} & \multicolumn{3}{c}{\textbf{WBCIC-MI-3C}} & \multicolumn{3}{c}{\textbf{WBCIC-MI-2C}} \\
\cmidrule(lr){2-4} \cmidrule(lr){5-7} \cmidrule(lr){8-10}
& Accuracy (\%) & Macro-F1 (\%) & Cohen's $\kappa$ & Accuracy (\%) & Macro-F1 (\%) & Cohen's $\kappa$ & Accuracy (\%) & Macro-F1 (\%) & Cohen's $\kappa$ \\
\midrule
BENDR & 52.23 $\pm$ 02.71\sig & 51.85 $\pm$ 02.76 & 0.045 $\pm$ 0.054 & 35.50 $\pm$ 02.76\sig & 35.39 $\pm$ 02.86 & 0.032 $\pm$ 0.041 & 51.09 $\pm$ 03.71\sig & 50.96 $\pm$ 03.81 & 0.022 $\pm$ 0.074 \\
BIOT & 52.83 $\pm$ 06.05\sig & 50.85 $\pm$ 07.72 & 0.057 $\pm$ 0.121 & 35.95 $\pm$ 04.36\sig & 33.23 $\pm$ 05.09 & 0.039 $\pm$ 0.065 & 50.83 $\pm$ 03.53\sig & 50.26 $\pm$ 03.75 & 0.017 $\pm$ 0.071 \\
LaBraM & 55.00 $\pm$ 06.56\sig & 53.65 $\pm$ 06.81 & 0.100 $\pm$ 0.131 & 39.89 $\pm$ 05.82\sig & 36.70 $\pm$ 06.77 & 0.098 $\pm$ 0.087 & 57.24 $\pm$ 05.97\sig & 56.51 $\pm$ 06.22 & 0.145 $\pm$ 0.119 \\
CBraMod & 71.16 $\pm$ 08.89 & 70.68 $\pm$ 09.44 & 0.423 $\pm$ 0.178 & 60.20 $\pm$ 10.76\sig & 59.23 $\pm$ 11.13 & 0.403 $\pm$ 0.161 & 78.14 $\pm$ 11.14\sig & 77.90 $\pm$ 11.43 & 0.563 $\pm$ 0.223 \\
CodeBrain & 65.09 $\pm$ 06.86\sig & 64.39 $\pm$ 07.24 & 0.302 $\pm$ 0.137 & 51.53 $\pm$ 09.54\sig & 49.53 $\pm$ 10.86 & 0.273 $\pm$ 0.143 & 74.17 $\pm$ 11.94\sig & 73.85 $\pm$ 12.23 & 0.483 $\pm$ 0.239 \\
CSBrain & 68.96 $\pm$ 11.94\sig & 68.61 $\pm$ 12.18 & 0.379 $\pm$ 0.239 & 63.05 $\pm$ 13.81\sig & 62.98 $\pm$ 13.85 & 0.446 $\pm$ 0.207 & 79.29 $\pm$ 12.76 & 79.19 $\pm$ 12.85 & 0.586 $\pm$ 0.255 \\
LUNA & 59.58 $\pm$ 07.03\sig & 59.16 $\pm$ 07.07 & 0.192 $\pm$ 0.141 & 45.35 $\pm$ 06.30\sig & 43.96 $\pm$ 06.60 & 0.180 $\pm$ 0.094 & 62.27 $\pm$ 09.00\sig & 61.91 $\pm$ 09.07 & 0.245 $\pm$ 0.180 \\
MIRepNet & 72.68 $\pm$ 17.40 & 72.49 $\pm$ 17.56 & 0.454 $\pm$ 0.348 & 48.28 $\pm$ 12.64\sig & 47.60 $\pm$ 13.11 & 0.224 $\pm$ 0.190 & 61.26 $\pm$ 10.87\sig & 60.83 $\pm$ 10.99 & 0.225 $\pm$ 0.217 \\
\midrule
CSP & 62.56 $\pm$ 14.27\sig & 58.58 $\pm$ 17.71 & 0.251 $\pm$ 0.285 & 34.77 $\pm$ 02.28\sig & 28.57 $\pm$ 05.71 & 0.022 $\pm$ 0.034 & 49.71 $\pm$ 03.66\sig & 44.98 $\pm$ 05.54 & -0.006 $\pm$ 0.073 \\
Welch PSD & 55.98 $\pm$ 07.40\sig & 53.52 $\pm$ 09.65 & 0.120 $\pm$ 0.148 & 37.65 $\pm$ 05.51\sig & 28.62 $\pm$ 11.27 & 0.065 $\pm$ 0.083 & 54.34 $\pm$ 05.03\sig & 51.40 $\pm$ 07.48 & 0.087 $\pm$ 0.101 \\
\midrule
\textbf{SingLEM} (primary) & \textbf{74.58 $\pm$ 08.15} & \textbf{74.45 $\pm$ 08.24} & \textbf{0.492 $\pm$ 0.163} & \textbf{68.14 $\pm$ 12.95} & \textbf{68.03 $\pm$ 12.99} & \textbf{0.522 $\pm$ 0.194} & \textbf{79.68 $\pm$ 13.42} & \textbf{79.60 $\pm$ 13.49} & \textbf{0.594 $\pm$ 0.268} \\
\midrule
\multicolumn{10}{l}{\textbf{SingLEM Ablation Variants}} \\
SingLEM (all 71 datasets)  & 74.52 $\pm$ 08.05 & 74.40 $\pm$ 08.09 & 0.490 $\pm$ 0.161 & 68.17 $\pm$ 12.46 & 68.02 $\pm$ 12.54 & 0.523 $\pm$ 0.187 & 79.48 $\pm$ 13.30 & 79.39 $\pm$ 13.38 & 0.590 $\pm$ 0.266 \\
SingLEM (w/o feature emb.) & 71.93 $\pm$ 08.51 & 71.76 $\pm$ 08.59 & 0.439 $\pm$ 0.170 & 66.08 $\pm$ 12.57 & 65.93 $\pm$ 12.66 & 0.491 $\pm$ 0.189 & 78.21 $\pm$ 13.81 & 78.11 $\pm$ 13.88 & 0.564 $\pm$ 0.276 \\
\bottomrule
\end{tabular}
\par\vspace{1pt}
\parbox{\textwidth}{%
\raggedright\scriptsize
* denotes significantly lower held-out-subject accuracy than SingLEM (primary) after Holm correction ($p_{\mathrm{Holm}}<0.05$).%
}
\end{table*}

\begin{table*}[!t]
\centering
\caption{Strict LOSO results on three cognitive task datasets (mean $\pm$ sd across held-out subjects). Best means among the primary comparison methods are bold; ablation variants are excluded from the ranking.}
\label{tab:Cognitive_results}
\scriptsize
\setlength{\tabcolsep}{3pt}
\begin{tabular}{@{}l ccc ccc ccc@{}}
\toprule
 
\multirow{2}{*}{\textbf{Model}} & \multicolumn{3}{c}{\textbf{N-back-2C}} & \multicolumn{3}{c}{\textbf{DSR-2C}} & \multicolumn{3}{c}{\textbf{WG-2C}} \\
\cmidrule(lr){2-4} \cmidrule(lr){5-7} \cmidrule(lr){8-10}
& Accuracy (\%) & Macro-F1 (\%) & Cohen's $\kappa$ & Accuracy  (\%) & Macro-F1 (\%) & Cohen's $\kappa$ & Accuracy (\%) & Macro-F1 (\%) & Cohen's $\kappa$ \\
\midrule
BENDR & 62.75 $\pm$ 05.69\sig & 62.19 $\pm$ 06.16 & 0.255 $\pm$ 0.114 & 59.62 $\pm$ 07.36\sig & 58.85 $\pm$ 07.89 & 0.192 $\pm$ 0.147 & 50.00 $\pm$ 09.13\sig & 49.21 $\pm$ 09.38 & 0.000 $\pm$ 0.183 \\
BIOT & 60.47 $\pm$ 09.78\sig & 57.99 $\pm$ 12.53 & 0.209 $\pm$ 0.196 & 61.70 $\pm$ 09.04\sig & 60.06 $\pm$ 09.81 & 0.234 $\pm$ 0.181 & 57.18 $\pm$ 07.86\sig & 55.15 $\pm$ 09.77 & 0.144 $\pm$ 0.157 \\
LaBraM & 62.25 $\pm$ 08.45\sig & 59.68 $\pm$ 11.48 & 0.245 $\pm$ 0.169 & 66.72 $\pm$ 11.96\sig & 65.50 $\pm$ 13.12 & 0.334 $\pm$ 0.239 & 61.54 $\pm$ 08.87\sig & 57.45 $\pm$ 12.82 & 0.231 $\pm$ 0.177 \\
CBraMod & 78.03 $\pm$ 09.98\sig & 76.63 $\pm$ 12.40 & 0.561 $\pm$ 0.200 & 79.38 $\pm$ 09.84\sig & 78.51 $\pm$ 10.94 & 0.588 $\pm$ 0.197 & 69.94 $\pm$ 08.35 & 68.86 $\pm$ 09.09 & 0.399 $\pm$ 0.167 \\
CodeBrain & 80.13 $\pm$ 10.33\sig & 79.16 $\pm$ 12.40 & 0.603 $\pm$ 0.207 & 82.00 $\pm$ 09.54\sig & 81.47 $\pm$ 10.22 & 0.640 $\pm$ 0.191 & 66.35 $\pm$ 09.65 & 63.60 $\pm$ 13.01 & 0.327 $\pm$ 0.193 \\
CSBrain & 78.13 $\pm$ 08.38\sig & 77.56 $\pm$ 08.85 & 0.563 $\pm$ 0.168 & 76.44 $\pm$ 10.92\sig & 75.67 $\pm$ 11.71 & 0.529 $\pm$ 0.218 & 68.01 $\pm$ 07.67 & 66.78 $\pm$ 08.79 & 0.360 $\pm$ 0.153 \\
LUNA & 69.37 $\pm$ 09.57\sig & 68.15 $\pm$ 10.46 & 0.387 $\pm$ 0.191 & 70.73 $\pm$ 08.81\sig & 69.39 $\pm$ 09.97 & 0.415 $\pm$ 0.176 & 58.59 $\pm$ 09.63\sig & 54.38 $\pm$ 12.45 & 0.172 $\pm$ 0.193 \\
\midrule
CSP & 58.37 $\pm$ 08.44\sig & 53.92 $\pm$ 11.89 & 0.167 $\pm$ 0.169 & 56.41 $\pm$ 08.56\sig & 49.06 $\pm$ 12.59 & 0.128 $\pm$ 0.171 & 53.53 $\pm$ 10.01\sig & 48.46 $\pm$ 13.09 & 0.071 $\pm$ 0.200 \\
Welch PSD & 64.78 $\pm$ 08.67\sig & 60.88 $\pm$ 12.37 & 0.296 $\pm$ 0.173 & 62.29 $\pm$ 09.53\sig & 56.90 $\pm$ 13.79 & 0.246 $\pm$ 0.191 & 61.47 $\pm$ 08.33\sig & 56.84 $\pm$ 13.03 & 0.229 $\pm$ 0.167 \\
\midrule
\textbf{SingLEM} (primary) & \textbf{84.15 $\pm$ 09.34} & \textbf{83.57 $\pm$ 11.32} & \textbf{0.683 $\pm$ 0.187} & \textbf{85.68 $\pm$ 08.75} & \textbf{85.46 $\pm$ 09.07} & \textbf{0.714 $\pm$ 0.175} & \textbf{70.26 $\pm$ 08.26} & \textbf{69.68 $\pm$ 08.70} & \textbf{0.405 $\pm$ 0.165} \\
\midrule
\multicolumn{10}{l}{\textbf{SingLEM Ablation Variants}} \\
SingLEM (all 71 datasets)  & 84.37 $\pm$ 09.77 & 83.71 $\pm$ 11.95 & 0.687 $\pm$ 0.195 & 85.90 $\pm$ 08.53 & 85.69 $\pm$ 08.84 & 0.718 $\pm$ 0.171 & 69.62 $\pm$ 08.68 & 69.23 $\pm$ 08.85 & 0.392 $\pm$ 0.174 \\
SingLEM (w/o feature emb.) & 83.65 $\pm$ 09.70 & 83.05 $\pm$ 11.66 & 0.673 $\pm$ 0.194 & 85.52 $\pm$ 08.86 & 85.20 $\pm$ 09.36 & 0.710 $\pm$ 0.177 & 69.55 $\pm$ 08.49 & 68.99 $\pm$ 08.84 & 0.391 $\pm$ 0.170 \\
\bottomrule
\end{tabular}
\par\vspace{1pt}
\parbox{\textwidth}{%
\raggedright\scriptsize
* denotes significantly lower held-out-subject accuracy than SingLEM (primary) after Holm correction ($p_{\mathrm{Holm}}<0.05$).%
}
\end{table*}

\subsection{Additional Classifier and Subject-Adapted Analyses}

To further assess the robustness, we evaluated feature-based models with MLP classifiers and compared them with supervised neural decoders. As shown in Table~\ref{tab:strict_mlp_neural_acc}, SingLEM achieved the best strict LOSO accuracy on five of the six datasets among the feature-based models using MLP classifiers. The only exception was \textbf{WG-2C}, where CBraMod achieved the highest accuracy. Compared with supervised neural decoders, SingLEM achieved higher accuracy than EEGNet, EEGConformer, and IFNetV2 on all six datasets.

The subject-adapted LOSO results are reported in Table~\ref{tab:adapted_acc}. With a small amount of target-subject calibration data, SingLEM achieved the best accuracy on all three MI datasets, \textbf{N-back-2C}, and \textbf{DSR-2C}. IFNetV2 achieved the highest accuracy on \textbf{WG-2C}. Overall, the adapted analysis shows that SingLEM remains strong under limited target-subject calibration, while the benefit of adaptation depends on the dataset and downstream model.

Across Tables~\ref{tab:MI_results}--\ref{tab:adapted_acc}, SingLEM achieved significantly higher held-out-subject accuracy than multiple baselines on every dataset, including all applicable baselines on WBCIC-MI-3C, N-back-2C, and DSR-2C in the primary SVM evaluation; no baseline significantly outperformed SingLEM in any setting.

\begin{table}[t]
\centering
\caption{Mean strict LOSO accuracy (\%) results using MLP classifiers and supervised neural models.}
\label{tab:strict_mlp_neural_acc}
\scriptsize
\setlength{\tabcolsep}{4pt}
\renewcommand{\arraystretch}{0.9}
\begin{tabular}{@{}l >{\centering\arraybackslash}p{0.85cm} >{\centering\arraybackslash}p{0.85cm} >{\centering\arraybackslash}p{0.85cm} >{\centering\arraybackslash}p{0.85cm} >{\centering\arraybackslash}p{0.85cm} >{\centering\arraybackslash}p{0.85cm}@{}}
\toprule
\textbf{Model}& \textbf{Dreyer-2C}& \textbf{WBCIC-3C}& \textbf{WBCIC-2C}& \textbf{N-back-2C}& \textbf{DSR-2C}& \textbf{WG-2C}\\
\midrule
\multicolumn{7}{l}{\textbf{EEG Foundation Models}} \\
\textbf{SingLEM} (primary)& \textbf{73.63}& \textbf{67.44}& \textbf{79.83}& \textbf{82.80}& \textbf{83.87}& 68.78\\
BENDR& 49.49\sig& 35.43\sig& 50.82\sig& 56.87\sig& 54.43\sig& 51.15\sig\\
BIOT& 50.18\sig& 35.62\sig& 49.82\sig& 58.12\sig& 59.56\sig& 53.53\sig\\
LaBraM& 54.67\sig& 37.56\sig& 51.59\sig& 59.69\sig& 60.63\sig& 61.09\\
CBraMod& 70.57& 60.81\sig& 78.49\sig& 75.85\sig& 77.78\sig& \textbf{70.45}\\
CodeBrain& 65.74\sig& 53.68\sig& 74.25\sig& 78.31\sig& 81.89& 67.18\\
CSBrain& 68.39& 64.20& 79.36& 77.56\sig& 75.80\sig& 65.38\\
LUNA& 57.23\sig& 41.80\sig& 59.56\sig& 63.28\sig& 64.64\sig& 59.68\sig\\
MIRepNet& 73.04& 48.71& 61.73\sig& --& --& --\\
\midrule
\multicolumn{7}{l}{\textbf{Classical EEG Features}} \\
CSP& 64.49& 35.01\sig& 50.12\sig& 57.94\sig& 57.64\sig& 54.74\sig\\
Welch PSD& 56.19\sig& 38.28\sig& 54.78\sig& 66.38\sig& 64.74\sig& 61.22\\
\midrule
\multicolumn{7}{l}{\textbf{Supervised Neural Decoders}} \\
EEGNet& 73.10& 60.53& 73.01\sig& 73.61\sig& 78.04\sig& 68.33\\
EEGConformer& 70.18& 51.99\sig& 71.91\sig& 77.71\sig& 77.62\sig& 60.38\sig\\
IFNetV2& 68.45& 54.38\sig& 73.94\sig& 72.04\sig& 73.66\sig& 67.56\\
\bottomrule
\end{tabular}
\par\vspace{1pt}
\parbox{\linewidth}{%
\raggedright\scriptsize
* denotes significantly lower held-out-subject accuracy than SingLEM (primary) after Holm correction ($p_{\mathrm{Holm}}<0.05$).%
}
\end{table}

\begin{table}[t]
\centering
\caption{Mean subject-adapted LOSO accuracy (\%) results using MLP classifiers and supervised neural models.}
\label{tab:adapted_acc}
\scriptsize
\setlength{\tabcolsep}{4pt}
\renewcommand{\arraystretch}{0.9}
\begin{tabular}{@{}l >{\centering\arraybackslash}p{0.85cm} >{\centering\arraybackslash}p{0.85cm} >{\centering\arraybackslash}p{0.85cm} >{\centering\arraybackslash}p{0.85cm} >{\centering\arraybackslash}p{0.85cm} >{\centering\arraybackslash}p{0.85cm}@{}}
\toprule
\textbf{Model}& \textbf{Dreyer-2C}& \textbf{WBCIC-3C}& \textbf{WBCIC-2C}& \textbf{N-back-2C}& \textbf{DSR-2C}& \textbf{WG-2C}\\
\midrule
\multicolumn{7}{l}{\textbf{EEG Foundation Models}} \\
\textbf{SingLEM} (primary)& \textbf{73.94}& \textbf{70.59}& \textbf{80.56}& \textbf{84.41}& \textbf{83.69}& 69.87\\
BENDR& 50.09\sig& 35.43\sig& 50.25\sig& 56.73\sig& 54.15\sig& 50.92\sig\\
BIOT& 50.34\sig& 36.99\sig& 49.72\sig& 60.48\sig& 59.62\sig& 54.76\sig\\
LaBraM& 54.63\sig& 39.02\sig& 52.34\sig& 62.10\sig& 60.54\sig& 60.62\sig\\
CBraMod& 71.17& 65.61& 79.45& 81.07& 80.54& 72.62\\
CodeBrain& 66.41\sig& 59.51\sig& 76.44\sig& 83.91& 82.31& 69.87\\
CSBrain& 70.79& 67.43& 79.86& 79.25\sig& 76.69\sig& 69.41\\
LUNA& 57.95\sig& 43.53\sig& 60.66\sig& 64.63\sig& 64.69\sig& 59.34\sig\\
MIRepNet& 73.51& 48.29& 61.64\sig& --& --& --\\
\midrule
\multicolumn{7}{l}{\textbf{Classical EEG Features}} \\
CSP& 65.05& 36.16\sig& 50.56\sig& 60.37\sig& 58.69\sig& 57.05\sig\\
Welch PSD& 59.31\sig& 42.53\sig& 55.39\sig& 77.02\sig& 69.77\sig& 71.43\\
\midrule
\multicolumn{7}{l}{\textbf{Supervised Neural Decoders}} \\
EEGNet& 72.66& 62.02& 76.24\sig& 74.34\sig& 77.77\sig& 68.86\\
EEGConformer& 70.54& 52.41\sig& 71.55\sig& 78.24\sig& 77.62\sig& 60.35\sig\\
IFNetV2& 70.92& 60.59\sig& 77.54\sig& 79.00& 77.69\sig& \textbf{74.18}\\
\bottomrule
\end{tabular}
\par\vspace{1pt}
\parbox{\linewidth}{%
\raggedright\scriptsize
* denotes significantly lower held-out-subject accuracy than SingLEM (primary) after Holm correction ($p_{\mathrm{Holm}}<0.05$).%
}
\end{table}

\subsection{Single-Channel Performance Analysis and Visualization}
\begin{figure*}[h!]
    \centering
    \includegraphics[width=1.0\textwidth]{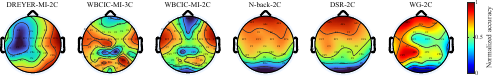} 
    \caption{Topographical accuracy maps for the Dreyer-MI-2C, WBCIC-MI-3C, WBCIC-MI-2C, N-back-2C, DSR-2C, and WG-2C datasets. Values were min--max normalized independently within each dataset; colors indicate relative within-dataset performance and are not directly comparable across datasets.}
    \label{fig:topo_maps}
\end{figure*}

A key advantage of SingLEM is its ability to evaluate the contribution of each electrode independently, providing insights into the spatial distribution of task-relevant neural information. We analyzed the single-channel classification performance across all six datasets, and the corresponding spatial accuracy patterns are visualized as topographical maps in Fig.~\ref{fig:topo_maps}. The single-channel analysis is intended primarily as a spatial interpretability analysis rather than as the main decoding setting. Individual electrodes provide only partial information, and the main decoding results therefore use channel-wise feature concatenation to integrate complementary representations when multiple electrodes are available. Complete per-electrode accuracy values underlying the topographical maps in Fig.~\ref{fig:topo_maps} are available in the project repository.

\subsubsection{MI Tasks}
Across all three MI datasets, high-performing electrodes were generally concentrated over scalp regions associated with motor-related processes. As shown in Fig.~\ref{fig:topo_maps}, for the \textbf{Dreyer-MI-2C} dataset, the highest performance was concentrated over right sensorimotor-related scalp regions, reflecting the contralateral organization of hand motor control. For the \textbf{WBCIC-MI} datasets, a similar pattern was observed, with higher performance obtained from fronto-temporal and central sites. Interestingly, on the \textbf{WBCIC-MI-2C} dataset, some posterior channels also showed above-average performance, suggesting a potential influence from non-motor cognitive processes such as visual cue processing.

\subsubsection{Cognitive Tasks} The analysis of the three cognitive tasks revealed a distinct spatial pattern, with high-performing electrodes primarily located over frontal scalp regions~\cite{cognitive-1, cognitive-2}. For both the \textbf{N-back} (working memory) and \textbf{DSR} (cognitive attention) tasks, the highest classification accuracies were overwhelmingly concentrated in the frontal regions. The \textbf{WG-2C} (cognitive load) task showed a more distributed pattern. While frontal areas remained important, the highest accuracies were also observed over the left central-parietal and parietal regions. This is neurophysiologically consistent with the linguistic and associative processing demands of the WG task, which engages a broader network beyond just the prefrontal region.

Collectively, these single-channel results suggest that SingLEM captures spatially meaningful and task-relevant EEG patterns.

\subsection{Ablation: Generalizability and Architectural Components}
\label{sec:ablation}
To assess the robustness of our pretraining strategy and the contribution of architectural components, we conducted two ablation studies. The corresponding variants are reported at the bottom of Tables~\ref{tab:MI_results} and~\ref{tab:Cognitive_results}.

First, we compared SingLEM (primary) with SingLEM (all 71 datasets). Mean accuracy differed by no more than 0.64 percentage points across the six datasets, suggesting limited dependence on prior exposure to the downstream source datasets.

Second, we evaluated the role of the feature embedding module using SingLEM (w/o feature emb.). For the shorter MI tasks (4--5 s), performance degraded notably, with accuracy dropping by up to 3 percentage points (e.g., Dreyer-MI-2C). In contrast, for the longer cognitive tasks (10 s), the impact was relatively small. This pattern highlights a functional interplay within the hierarchical design: the feature embedding module supplies local context by capturing short-range temporal dependencies, while the global transformer encoder alone suffices for modeling longer sequences.

These findings validate both the generalizability of SingLEM’s pretraining and the importance of its hierarchical architecture, particularly for tasks involving short trial durations.

\section{Discussion}
\label{sec:discussion}
 
This study introduced SingLEM, a self-supervised representation learning framework based on a single-channel EEG encoder. The results highlight four main findings. First, SingLEM achieved the best performance among the primary comparison methods across all six datasets under the primary strict LOSO SVM evaluation. Second, it remained competitive under alternative classifiers and partially subject-adapted settings, although the best-performing model depended on the downstream configuration. Third, its channel-wise representations enabled spatial analysis of task-discriminative EEG information across electrodes. Fourth, the ablation results indicated limited dependence on prior exposure to downstream datasets and a greater contribution of the feature embedding module on the evaluated MI datasets.

\subsection{Performance and Methodological Implications}
 
In the primary strict LOSO SVM evaluation, SingLEM achieved the best performance among the primary comparison methods across all six downstream datasets, suggesting that pretraining on diverse single-channel EEG signals can yield transferable representations for unseen subjects without downstream fine-tuning of the encoder. Although each channel is encoded independently, concatenating the resulting channel-wise representations provided an effective feature-level aggregation strategy for multi-channel decoding, thereby allowing downstream classifiers to exploit spatially distributed information.

The additional MLP, supervised neural decoder, and partially subject-adapted analyses showed that SingLEM remained competitive across alternative downstream settings, although the best-performing model varied by dataset, classifier, and adaptation protocol. Thus, SingLEM should be viewed as a strong frozen representation learner rather than as universally dominant across all downstream configurations. Its use of frozen features with lightweight classifiers also offers practical value when target-subject calibration data or computational resources are limited.

The comparison with other pretrained EEG models shows that frozen-feature evaluation is distinct from model-specific supervised adaptation. The weaker frozen-feature performance of some pretrained models does not necessarily imply that these models are ineffective but may reflect differences in architectural design, channel-format requirements, or the need for model-specific adaptation or fine-tuning. In contrast, SingLEM was designed to provide channel-wise representations that can be directly reused across different electrode layouts.

The additional SingLEM configurations reported in Tables~\ref{tab:MI_results} and~\ref{tab:Cognitive_results} further support these conclusions. Pretraining on all 71 datasets produced only minor differences from SingLEM (primary), indicating limited dependence on prior exposure to the evaluated datasets. In addition, removing the feature embedding module affected the shorter MI trials more than the longer cognitive trials, suggesting that the hierarchical architecture may help capture local temporal structure, especially when the available trial duration is short.

\subsection{Neurophysiological Plausibility and Paradigm Evaluation}
The single-channel analysis in this study not only quantified the performance of SingLEM but also provided supporting evidence that its learned representations reflect task-relevant EEG patterns. This trend was observed across both motor and cognitive domains.

For the MI tasks, high-performing channels were concentrated over sensorimotor-related scalp regions, including central, fronto-central, and fronto-temporal areas, which are commonly associated with motor planning and execution. The lateralized patterns observed for the Dreyer-MI dataset were also consistent with the contralateral organization of the motor system, supporting the neurophysiological plausibility of the learned representations.

For cognitive tasks, discriminative patterns shifted in accordance with task demands. In the N-back (working memory) and DSR (attention) tasks, performance was dominated by frontal electrodes, consistent with the role of prefrontal regions in attention-related and working-memory processes~\cite{cognitive-1, cognitive-2}. The WG task showed a broader distribution, with high performance in frontal regions as well as left central-parietal areas. This spatial pattern suggests that SingLEM captured task-relevant EEG information associated with linguistic and associative processing beyond general executive control.

Importantly, this channel-level granularity also supported the evaluation of experimental paradigms. In the WBCIC-MI-2C data, relatively high performance was observed in occipital and parietal regions, possibly reflecting visual cue processing in addition to MI-related activity. Such findings suggest that SingLEM can help identify spatial patterns that may reflect task-relevant information or possible confounding factors, providing an interpretability advantage over less transparent multi-channel models.

\subsection{Limitations and Future Work}
 
While this study demonstrates the potential of a channel-wise EEG representation learning framework, several limitations remain. First, although SingLEM was pretrained on a large and heterogeneous EEG corpus, the downstream evaluation focused on MI and cognitive classification tasks from three source studies. Further evaluation on other EEG applications, such as sleep staging, seizure detection, clinical abnormality detection, and emotion recognition, is needed to assess the generality of the learned representations. Second, the experiments used offline preprocessing, including zero-phase filtering, for consistent benchmarking. Online BCI deployment would require causal filtering or streaming-compatible preprocessing, which should be validated in future real-time studies.
 
Future work should also explore several methodological extensions. The current approach concatenates channel-wise representations before classification and was effective in the present experiments, but it does not explicitly model spatial relationships among electrodes. More advanced spatial aggregation methods could therefore further improve performance. In addition, although SingLEM is flexible with respect to channel layout, its representations may still be affected by reference-electrode differences, making reference-robust representation learning an important direction. Finally, beyond its use as a frozen EEG feature extractor, future work should investigate broader adaptation and continual learning strategies, including subject, device, and site adaptation, continual pretraining with new large-scale EEG datasets, and exploratory transfer to related biosignals or neurophysiological recordings such as electrocardiography, electromyography, or electrocorticography.

\section{Conclusion}
\label{sec:conclusion}
 
This study introduced SingLEM, a foundation model designed to reduce dependence on fixed electrode layouts by learning representations at the single-channel level. SingLEM (all 71 datasets) is provided for general reuse, whereas the leakage-controlled evaluation used SingLEM (primary), pretrained on 68 datasets after excluding the three downstream datasets.
   
The primary strict LOSO evaluation showed that SingLEM (primary) achieved the best performance among the primary comparison methods across all six downstream datasets when used as a frozen feature extractor with an SVM classifier. Complementary analyses with alternative classifiers and limited target-subject calibration further supported the robustness of the learned representations across downstream settings without encoder fine-tuning. SingLEM's channel-wise representations can be combined through channel-wise feature concatenation for multi-channel downstream decoding, while its single-channel granularity supports spatial analysis of task-relevant EEG patterns by evaluating individual electrodes. By decoupling feature learning from fixed multi-channel input layouts, SingLEM provides a flexible representation-learning framework across diverse electrode configurations.

\section*{Acknowledgment}
We would like to thank Editage (\url{www.editage.jp}) for English language editing.

\bibliographystyle{IEEEtran}
\bibliography{bibs}

@IEEEtranBSTCTL{IEEEtranBSTCTL,
  CTLuse_forced_etal       = "yes",
  CTLmax_names_forced_etal = "1",
  CTLnames_show_etal       = "1",
  CTLjournal_names         = "abbreviated",
  CTLlines_around_url      = "yes"
}

@inproceedings{devlin2019bert,
  title     = {{{BERT}}: Pre-training of deep bidirectional transformers for language understanding},
  author    = {Devlin, Jacob and Chang, Ming-Wei and Lee, Kenton and Toutanova, Kristina},
  booktitle = {Proc. Conf. North Amer. Chapter Assoc. Comput. Linguistics},
  address   = {Minneapolis, MN, USA},
  pages     = {4171--4186},
  year      = {2019}
}

@article{achiam2023gpt,
  title   = {{{GPT-4}} technical report},
  author  = {Achiam, Josh and Adler, Steven and Agarwal, Sandhini and Ahmad, Lama and Akkaya, Ilge and Aleman, Florencia Leoni and Almeida, Diogo and Altenschmidt, Janko and Altman, Sam and Anadkat, Shyamal and et al.},
  journal = {arXiv:2303.08774},
  year    = {2023}
}

@article{Foundation_model_review,
  author  = {He, Yuting and Huang, Fuxiang and Jiang, Xinrui and Nie, Yuxiang and Wang, Minghao and Wang, Jiguang and Chen, Hao},
  journal = {IEEE Rev. Biomed. Eng.},
  title   = {Foundation model for advancing healthcare: Challenges, opportunities and future directions},
  year    = {2025},
  volume  = {18},
  pages   = {172--191},
  doi     = {10.1109/RBME.2024.3496744}
}

@article{FM_biomedical_eng,
  author  = {Qiu, Jianing and Li, Lin and Sun, Jiankai and Peng, Jiachuan and Shi, Peilun and Zhang, Ruiyang and Dong, Yinzhao and Lam, Kyle and Lo, Frank P.-W. and Xiao, Bo and Yuan, Wu and Wang, Ningli and Xu, Dong and Lo, Benny},
  journal = {IEEE J. Biomed. Health Inform.},
  title   = {Large {AI} models in health informatics: Applications, challenges, and the future},
  year    = {2023},
  volume  = {27},
  number  = {12},
  pages   = {6074--6087},
  doi     = {10.1109/JBHI.2023.3316750}
}

@inproceedings{he2022masked,
  title     = {Masked autoencoders are scalable vision learners},
  author    = {He, Kaiming and Chen, Xinlei and Xie, Saining and Li, Yanghao and Doll{\'a}r, Piotr and Girshick, Ross},
  booktitle = {Proc. IEEE/CVF Conf. Comput. Vis. Pattern Recognit.},
  address   = {New Orleans, LA, USA},
  pages     = {16000--16009},
  year      = {2022}
}

@article{MI-BCI-popular-1,
  author  = {Yuan, Han and He, Bin},
  journal = {IEEE Trans. Biomed. Eng.},
  title   = {Brain--computer interfaces using sensorimotor rhythms: Current state and future perspectives},
  year    = {2014},
  volume  = {61},
  number  = {5},
  pages   = {1425--1435},
  doi     = {10.1109/TBME.2014.2312397}
}

@article{MI-BCI-recent-1,
  author  = {Chen, Peiyin and Liu, Xiaofeng and Ma, Chao and Wang, He and Yang, Xiong and Grebogi, Celso and Gu, Xiao and Gao, Zhongke},
  journal = {IEEE J. Biomed. Health Inform.},
  title   = {Unsupervised domain adaptation with synchronized self-training for cross-domain motor imagery recognition},
  year    = {2025},
  volume  = {29},
  number  = {5},
  pages   = {3664--3677},
  doi     = {10.1109/JBHI.2025.3525577}
}

@article{MI-BCI-recent-2,
  author  = {Liu, Guoyang and Zhang, Rui and Tian, Lan and Zhou, Weidong},
  journal = {IEEE J. Biomed. Health Inform.},
  title   = {Fine-grained spatial-frequency-time framework for motor imagery brain--computer interface},
  year    = {2025},
  volume  = {29},
  number  = {6},
  pages   = {4121--4133},
  doi     = {10.1109/JBHI.2025.3536212}
}

@article{Sleep-popular-1,
  author  = {Supratak, Akara and Dong, Hao and Wu, Chao and Guo, Yike},
  journal = {IEEE Trans. Neural Syst. Rehabil. Eng.},
  title   = {{DeepSleepNet}: A model for automatic sleep stage scoring based on raw single-channel {{EEG}}},
  year    = {2017},
  volume  = {25},
  number  = {11},
  pages   = {1998--2008},
  doi     = {10.1109/TNSRE.2017.2721116}
}

@article{Sleep-1,
  author  = {Wan, Cheng and Nnamdi, Micky C. and Shi, Wenqi and Smith, Benjamin and Purnell, Chad and Wang, May D.},
  journal = {IEEE J. Biomed. Health Inform.},
  title   = {Advancing sleep disorder diagnostics: A transformer-based {{EEG}} model for sleep stage classification and {{OSA}} prediction},
  year    = {2025},
  volume  = {29},
  number  = {2},
  pages   = {878--886},
  doi     = {10.1109/JBHI.2024.3512616}
}

@article{Seizure-shoji2021automated,
  title   = {Automated detection of abnormalities from an {{EEG}} recording of epilepsy patients with a compact convolutional neural network},
  author  = {Shoji, Taku and Yoshida, Noboru and Tanaka, Toshihisa},
  journal = {Biomed. Signal Process. Control},
  volume  = {70},
  pages   = {103013},
  year    = {2021}
}

@article{Seizure-2,
  author  = {Meng, Kunying and Wang, Denghai and Zhang, Donghui and Guo, Kunlin and Lu, Kai and Lu, Junfeng and Yu, Renping and Zhang, Lipeng and Hu, Yuxia and Zhang, Rui and Chen, Mingming},
  journal = {IEEE J. Biomed. Health Inform.},
  title   = {Real-time epileptic seizure prediction method with spatio-temporal information transfer learning},
  year    = {2025},
  volume  = {29},
  number  = {3},
  pages   = {2222--2232},
  doi     = {10.1109/JBHI.2024.3509959}
}

@article{Diagnosis-review-1,
  author  = {Emad-Ud-Din, Muhammad and Almuteb, Ibrahim and Wang, Ya and Hubbard, James E.},
  journal = {IEEE Sensors J.},
  title   = {Electroencephalography machine-learning features and methods for early diagnosis and classification of {Parkinson’s} disease (2013–2023): A review},
  year    = {2025},
  volume  = {25},
  number  = {12},
  pages   = {21017--21032},
  doi     = {10.1109/JSEN.2025.3562448}
}

@article{Emotion-popular-1,
  author  = {Alarcão, Soraia M. and Fonseca, Manuel J.},
  journal = {IEEE Trans. Affect. Comput.},
  title   = {Emotions recognition using {{EEG}} signals: A survey},
  year    = {2019},
  volume  = {10},
  number  = {3},
  pages   = {374--393},
  doi     = {10.1109/TAFFC.2017.2714671}
}

@article{Emotion-recent-1,
  author  = {Ding, Yi and Tong, Chengxuan and Zhang, Shuailei and Jiang, Muyun and Li, Yong and Lim, Kevin JunLiang and Guan, Cuntai},
  journal = {IEEE Trans. Neural Netw. Learn. Syst.},
  title   = {{EmT}: A novel transformer for generalized cross-subject {{EEG}} emotion recognition},
  year    = {2025},
  volume  = {36},
  number  = {6},
  pages   = {10381--10393},
  doi     = {10.1109/TNNLS.2025.3552603}
}

@article{DNN-for-EEG-review1,
  title   = {Deep learning-based electroencephalography analysis: a systematic review},
  author  = {Roy, Yannick and Banville, Hubert and Albuquerque, Isabela and Gramfort, Alexandre and Falk, Tiago H and Faubert, Jocelyn},
  journal = {J. Neural Eng.},
  volume  = {16},
  number  = {5},
  pages   = {051001},
  year    = {2019}
}

@article{EEG-challenges,
  title   = {A review of classification algorithms for {{EEG}}-based brain--computer interfaces: a 10 year update},
  author  = {Lotte, Fabien and Bougrain, Laurent and Cichocki, Andrzej and Clerc, Maureen and Congedo, Marco and Rakotomamonjy, Alain and Yger, Florian},
  journal = {J. Neural Eng.},
  volume  = {15},
  number  = {3},
  pages   = {031005},
  year    = {2018}
}

@article{SSVEP-BCI-1,
  author  = {He, Xinjie and Allison, Brendan Z. and Liang, Wei and Chen, Weijie and Wang, Xingyu and Cichocki, Andrzej and Jin, Jing},
  journal = {IEEE Sensors J.},
  title   = {Leveraging peripheral visual stimuli for enhanced {{SSVEP}}-based {{BCI}}s in fast calibration scenario},
  year    = {2025},
  volume  = {25},
  number  = {10},
  pages   = {17683--17695},
  doi     = {10.1109/JSEN.2025.3554985}
}

@article{ViT,
  title   = {An image is worth 16x16 words: Transformers for image recognition at scale},
  author  = {Dosovitskiy, Alexey and Beyer, Lucas and Kolesnikov, Alexander and Weissenborn, Dirk and Zhai, Xiaohua and Unterthiner, Thomas and Dehghani, Mostafa and Minderer, Matthias and Heigold, Georg and Gelly, Sylvain and et al.},
  journal = {arXiv:2010.11929},
  year    = {2020}
}

@article{BENDR,
  title   = {{{BENDR}}: Using transformers and a contrastive self-supervised learning task to learn from massive amounts of {{EEG}} data},
  author  = {Kostas, Demetres and Aroca-Ouellette, Stephane and Rudzicz, Frank},
  journal = {Front. Hum. Neurosci.},
  volume  = {15},
  pages   = {653659},
  year    = {2021}
}

@inproceedings{BIOT,
  title   = {{BIOT}: Biosignal transformer for cross-data learning in the wild},
  author  = {Yang, Chaoqi and Westover, M and Sun, Jimeng},
  booktitle = {Adv. Neural Inf. Process. Syst.},
  volume  = {36},
  pages   = {78240--78260},
  address = {New Orleans, LA, USA},
  year    = {2023}
}

@inproceedings{LaBraM,
  title     = {Large brain model for learning generic representations with tremendous {{EEG}} data in {{BCI}}},
  author    = {Jiang, Weibang and Zhao, Liming and Lu, Bao-liang},
  booktitle = {Proc. Int. Conf. Learn. Represent.},
  address   = {Vienna, Austria},
  pages     = {16405--16426},
  year      = {2024}
}

@inproceedings{CBraMod,
  title     = {{{CBraMod}}: A criss-cross brain foundation model for {{EEG}} decoding},
  author    = {Wang, Jiquan and Zhao, Sha and Luo, Zhiling and Zhou, Yangxuan and Jiang, Haiteng and Li, Shijian and Li, Tao and Pan, Gang},
  booktitle = {Proc. Int. Conf. Learn. Represent.},
  address   = {Singapore},
  pages     = {75310--75346},
  year      = {2025}
}

@inproceedings{CodeBrain,
  title     = {{CodeBrain}: Bridging decoupled tokenizer and multi-scale architecture for {{EEG}} foundation model},
  author    = {Ma, Jingying and Wu, Feng and Lin, Qika and Xing, Yucheng and Liu, Chenyu and Jia, Ziyu and Feng, Mengling},
  booktitle = {Proc. Int. Conf. Learn. Represent.},
  address   = {Rio de Janeiro, Brazil},
  year      = {2026}
}

@inproceedings{CSBrain,
  title     = {{CSBrain}: A cross-scale spatiotemporal brain foundation model for {{EEG}} decoding},
  author    = {Zhou, Yuchen and Wu, Jiamin and Ren, Zichen and Yao, Zhouheng and Lu, Weiheng and Peng, Kunyu and Zheng, Qihao and Song, Chunfeng and Ouyang, Wanli and Gou, Chao},
  booktitle = {Adv. Neural Inf. Process. Syst.},
  address   = {San Diego, CA, USA},
  pages     = {87150--87195},
  year      = {2025}
}

@inproceedings{LUNA,
  title     = {{LUNA}: Efficient and topology-agnostic foundation model for {{EEG}} signal analysis},
  author    = {D{\"o}ner, Berkay and Ingolfsson, Thorir Mar and Benini, Luca and Li, Yawei},
  booktitle = {Adv. Neural Inf. Process. Syst.},
  address   = {San Diego, CA, USA},
  pages     = {70682--70708},
  year      = {2025}
}

@article{MIRepNet,
  title   = {{MIRepNet}: A pipeline and pre-trained model for {EEG}-based motor imagery classification},
  author  = {Liu, Dingkun and Chen, Zhu and Luo, Jingwei and Lian, Shijie and Chen, Yuheng and Hou, Shaojie and Zhu, Xiaolian and Wu, Dongrui},
  journal = {Knowl.-Based Syst.},
  volume  = {343},
  pages   = {115966},
  year    = {2026}
}

@article{IFNet,
  title   = {{IFNet}: An interactive frequency convolutional neural network for enhancing motor imagery decoding from {EEG}},
  author  = {Wang, Jiaheng and Yao, Li and Wang, Yijun},
  journal = {IEEE Trans. Neural Syst. Rehabil. Eng.},
  volume  = {31},
  pages   = {1900--1911},
  year    = {2023}
}

@article{Conformer,
  author  = {Song, Yonghao and Zheng, Qingqing and Liu, Bingchuan and Gao, Xiaorong},
  journal = {IEEE Trans. Neural Syst. Rehabil. Eng.},
  title   = {{{EEG}} conformer: Convolutional transformer for {{EEG}} decoding and visualization},
  year    = {2023},
  volume  = {31},
  pages   = {710--719},
  doi     = {10.1109/TNSRE.2022.3230250}
}

@article{Deformer,
  author  = {Ding, Yi and Li, Yong and Sun, Hao and Liu, Rui and Tong, Chengxuan and Liu, Chenyu and Zhou, Xinliang and Guan, Cuntai},
  journal = {IEEE J. Biomed. Health Inform.},
  title   = {{{EEG-Deformer}}: A dense convolutional transformer for brain-computer interfaces},
  year    = {2025},
  volume  = {29},
  number  = {3},
  pages   = {1909--1918},
  doi     = {10.1109/JBHI.2024.3504604}
}

@article{EEGNet,
  title   = {{EEGNet}: a compact convolutional neural network for {{EEG}}-based brain--computer interfaces},
  author  = {Lawhern, Vernon J and Solon, Amelia J and Waytowich, Nicholas R and Gordon, Stephen M and Hung, Chou P and Lance, Brent J},
  journal = {J. Neural Eng.},
  volume  = {15},
  number  = {5},
  pages   = {056013},
  year    = {2018}
}

@article{deep-convnet,
  title   = {Deep learning with convolutional neural networks for {{EEG}} decoding and visualization},
  author  = {Schirrmeister, Robin Tibor and Springenberg, Jost Tobias and Fiederer, Lukas Dominique Josef and Glasstetter, Martin and Eggensperger, Katharina and Tangermann, Michael and Hutter, Frank and Burgard, Wolfram and Ball, Tonio},
  journal = {Hum. Brain Mapp.},
  volume  = {38},
  number  = {11},
  pages   = {5391--5420},
  year    = {2017}
}

@article{SSVEPformer,
  title   = {A transformer-based deep neural network model for {{SSVEP}} classification},
  author  = {Chen, Jianbo and Zhang, Yangsong and Pan, Yudong and Xu, Peng and Guan, Cuntai},
  journal = {Neural Netw.},
  volume  = {164},
  pages   = {521--534},
  year    = {2023}
}

@ARTICLE{GMAEEG,
  author={Fu, Zanhao and Zhu, Huaiyu and Zhao, Yisheng and Huan, Ruohong and Zhang, Yi and Chen, Shuohui and Pan, Yun},
  journal={IEEE J. Biomed. Health Inform.}, 
  title={{GMAEEG}: A Self-Supervised Graph Masked Autoencoder for {EEG} Representation Learning}, 
  year={2024},
  volume={28},
  number={11},
  pages={6486-6497},
  doi={10.1109/JBHI.2024.3443651}}

@ARTICLE{AEFBCSP,
  author={Mammone, Nadia and Ieracitano, Cosimo and Adeli, Hojjat and Morabito, Francesco C.},
  journal={IEEE J. Biomed. Health Inform.}, 
  title={AutoEncoder Filter Bank Common Spatial Patterns to Decode Motor Imagery From {EEG}}, 
  year={2023},
  volume={27},
  number={5},
  pages={2365-2376},
  doi={10.1109/JBHI.2023.3243698}}

@article{cho-h,
  title   = {{{EEG}} datasets for motor imagery brain--computer interface},
  author  = {Cho, Hohyun and Ahn, Minkyu and Ahn, Sangtae and Kwon, Moonyoung and Jun, Sung Chan},
  journal = {GigaScience},
  volume  = {6},
  number  = {7},
  pages   = {gix034},
  year    = {2017}
}

@article{schalk-g,
  author  = {Schalk, Gerwin},
  title   = {{EEG Motor Movement/Imagery Dataset}},
  journal = {PhysioNet},
  year    = {2009},
  note    = {Version 1.0.0, doi: 10.13026/C28G6P}
}

@article{shin-j-3,
  title   = {Open access dataset for {{EEG}}+{{NIRS}} single-trial classification},
  author  = {Shin, Jaeyoung and von L{\"u}hmann, Alexander and Blankertz, Benjamin and Kim, Do-Won and Jeong, Jichai and Hwang, Han-Jeong and M{\"u}ller, Klaus-Robert},
  journal = {IEEE Trans. Neural Syst. Rehabil. Eng.},
  volume  = {25},
  number  = {10},
  pages   = {1735--1745},
  year    = {2017}
}

@article{kaya-m,
  title   = {A large electroencephalographic motor imagery dataset for electroencephalographic brain computer interfaces},
  author  = {Kaya, Murat and Binli, Mustafa Kemal and Ozbay, Erkan and Yanar, Hilmi and Mishchenko, Yuriy},
  journal = {Sci. Data},
  volume  = {5},
  pages   = {180211},
  year    = {2018},
  doi     = {10.1038/sdata.2018.211}
}

@article{lee-m,
  title   = {{{EEG}} dataset and {OpenBMI} toolbox for three {{BCI}} paradigms: An investigation into {{BCI}} illiteracy},
  author  = {Lee, Min-Ho and Kwon, O-Yeon and Kim, Yong-Jeong and Kim, Hong-Kyung and Lee, Young-Eun and Williamson, John and Fazli, Siamac and Lee, Seong-Whan},
  journal = {GigaScience},
  volume  = {8},
  number  = {5},
  pages   = {giz002},
  year    = {2019}
}

@article{brandl-s,
  title   = {Motor imagery under distraction---An open access {{BCI}} dataset},
  author  = {Brandl, Stephanie and Blankertz, Benjamin},
  journal = {Front. Neurosci.},
  volume  = {14},
  pages   = {566147},
  year    = {2020}
}

@article{chen-z,
  title   = {Open access dataset integrating {{EEG}} and {fNIRS} during {Stroop} tasks},
  author  = {Chen, Zemeng and Gao, Chenyang and Li, Ting and Ji, Xiang and Liu, Shuyu and Xiao, Ming},
  journal = {Sci. Data},
  volume  = {10},
  pages   = {618},
  year    = {2023}
}

@article{mou-x,
  title   = {{ChineseEEG}: A {Chinese} linguistic corpora {{EEG}} dataset for semantic alignment and neural decoding},
  author  = {Mou, Xinyu and He, Cuilin and Tan, Liwei and Yu, Junjie and Liang, Huadong and Zhang, Jianyu and Tian, Yan and Yang, Yu-Fang and Xu, Ting and Wang, Qing and et al.},
  journal = {Sci. Data},
  volume  = {11},
  pages   = {550},
  year    = {2024}
}

@article{getzmann-s,
  title   = {Resting-state {{EEG}} data before and after cognitive activity across the adult lifespan and a 5-year follow-up},
  author  = {Getzmann, Stephan and Gajewski, Patrick D and Schneider, Daniel and Wascher, Edmund},
  journal = {Sci. Data},
  volume  = {11},
  pages   = {988},
  year    = {2024}
}

@article{ji-x,
  title   = {{{EEG}} and {fNIRS} datasets based on {Stroop} task during two weeks of high-altitude exposure in new immigrants},
  author  = {Ji, Xiang and Bao, Botao and Li, Lin Z and Pu, Jiangbo and Lin, Yu and Zhang, Xin and Chen, Zemeng and Li, Ting},
  journal = {Sci. Data},
  volume  = {11},
  pages   = {350},
  year    = {2024}
}

@article{momenian-m,
  title   = {Le petit prince hong kong (lpphk): Naturalistic {fMRI} and {{EEG}} data from older {Cantonese} speakers},
  author  = {Momenian, Mohammad and Ma, Zhengwu and Wu, Shuyi and Wang, Chengcheng and Brennan, Jonathan and Hale, John and Meyer, Lars and Li, Jixing},
  journal = {Sci. Data},
  volume  = {11},
  pages   = {992},
  year    = {2024}
}

@article{xiang-c,
  title   = {A resting-state {{EEG}} dataset for sleep deprivation},
  author  = {Xiang, Chuqin and Fan, Xinrui and Bai, Duo and Lv, Ke and Lei, Xu},
  journal = {Sci. Data},
  volume  = {11},
  pages   = {427},
  year    = {2024}
}

@article{babayan-a,
  author  = {Babayan, Anahit and Erbey, Miray and Kumral, Deniz and Reinelt, Janis D. and Reiter, Andrea M. F. and R{\"o}bbig, Josefin and Schaare, H. Lina and Uhlig, Marie and Anwander, Alfred and Bazin, Pierre-Louis and others},
  title   = {A mind-brain-body dataset of {MRI}, {{EEG}}, cognition, emotion, and peripheral physiology in young and old adults},
  journal = {Sci. Data},
  volume  = {6},
  pages   = {180308},
  year    = {2019},
  doi     = {10.1038/sdata.2018.308}
}

@article{van-dijk,
  title   = {The two decades brainclinics research archive for insights in neurophysiology ({TDBRAIN}) database},
  author  = {van Dijk, Hanneke and Van Wingen, Guido and Denys, Damiaan and Olbrich, Sebastian and Van Ruth, Rosalinde and Arns, Martijn},
  journal = {Sci. Data},
  volume  = {9},
  pages   = {333},
  year    = {2022}
}

@article{ngo-t,
  title   = {An {{EEG}} \& eye-tracking dataset of {ALS} patients \& healthy people during eye-tracking-based spelling system usage},
  author  = {Ngo, Thi Duyen and Kieu, Hai Dang and Nguyen, Minh Hoa and Nguyen, The Hoang-Anh and Can, Van Mao and Nguyen, Ba Hung and Le, Thanh Ha},
  journal = {Sci. Data},
  volume  = {11},
  pages   = {664},
  year    = {2024}
}

@article{grootswagers-t-16,
  title   = {Human {{EEG}} recordings for 1,854 concepts presented in rapid serial visual presentation streams},
  author  = {Grootswagers, Tijl and Zhou, Ivy and Robinson, Amanda K and Hebart, Martin N and Carlson, Thomas A},
  journal = {Sci. Data},
  volume  = {9},
  pages   = {3},
  year    = {2022}
}

@article{dzianok-p,
  title   = {{PEARL-Neuro} Database: {{EEG}}, {fMRI}, health and lifestyle data of middle-aged people at risk of dementia},
  author  = {Dzianok, Patrycja and Kublik, Ewa},
  journal = {Sci. Data},
  volume  = {11},
  pages   = {276},
  year    = {2024}
}

@article{ma-j,
  title   = {A large {{EEG}} dataset for studying cross-session variability in motor imagery brain-computer interface},
  author  = {Ma, Jun and Yang, Banghua and Qiu, Wenzheng and Li, Yunzhe and Gao, Shouwei and Xia, Xinxing},
  journal = {Sci. Data},
  volume  = {9},
  pages   = {531},
  year    = {2022}
}

@article{chen-y,
  title   = {An {{EEG}} dataset of neural signatures in a competitive two-player game encouraging deceptive behavior},
  author  = {Chen, Yiyu and Fazli, Siamac and Wallraven, Christian},
  journal = {Sci. Data},
  volume  = {11},
  pages   = {389},
  year    = {2024}
}

@article{cao-z,
  title   = {Multi-channel {{EEG}} recordings during a sustained-attention driving task},
  author  = {Cao, Zehong and Chuang, Chun-Hsiang and King, Jung-Kai and Lin, Chin-Teng},
  journal = {Sci. Data},
  volume  = {6},
  pages   = {19},
  year    = {2019}
}

@article{telesford-q,
  title   = {An open-access dataset of naturalistic viewing using simultaneous {{EEG}}-{fMRI}},
  author  = {Telesford, Qawi K and Gonzalez-Moreira, Eduardo and Xu, Ting and Tian, Yiwen and Colcombe, Stanley J and Cloud, Jessica and Russ, Brian E and Falchier, Arnaud and Nentwich, Maximilian and Madsen, Jens and et al.},
  journal = {Sci. Data},
  volume  = {10},
  pages   = {554},
  year    = {2023}
}

@article{ma-x,
  title   = {Multi-channel {{EEG}} recording during motor imagery of different joints from the same limb},
  author  = {Ma, Xuelin and Qiu, Shuang and He, Huiguang},
  journal = {Sci. Data},
  volume  = {7},
  pages   = {191},
  year    = {2020}
}

@article{yang-b,
  title   = {A multi-day and high-quality {{EEG}} dataset for motor imagery brain-computer interface},
  author  = {Yang, Banghua and Rong, Fenqi and Xie, Yunlong and Li, Du and Zhang, Jiayang and Li, Fu and Shi, Guangming and Gao, Xiaorong},
  journal = {Sci. Data},
  volume  = {12},
  pages   = {488},
  year    = {2025}
}

@article{chen-k,
  title   = {A resource for assessing dynamic binary choices in the adult brain using {{EEG}} and mouse-tracking},
  author  = {Chen, Kun and Wang, Ruien and Huang, Jiamin and Gao, Fei and Yuan, Zhen and Qi, Yanyan and Wu, Haiyan},
  journal = {Sci. Data},
  volume  = {9},
  pages   = {416},
  year    = {2022}
}

@article{dreyer-p,
  title   = {A large {{EEG}} database with users’ profile information for motor imagery brain-computer interface research},
  author  = {Dreyer, Pauline and Roc, Aline and Pillette, L{\'e}a and Rimbert, S{\'e}bastien and Lotte, Fabien},
  journal = {Sci. Data},
  volume  = {10},
  pages   = {580},
  year    = {2023}
}

@article{wang-y,
  title   = {A test-retest resting, and cognitive state {{EEG}} dataset during multiple subject-driven states},
  author  = {Wang, Yulin and Duan, Wei and Dong, Debo and Ding, Lihong and Lei, Xu},
  journal = {Sci. Data},
  volume  = {9},
  pages   = {566},
  year    = {2022}
}

@article{chen-j,
  title   = {A large finer-grained affective computing {{EEG}} dataset},
  author  = {Chen, Jingjing and Wang, Xiaobin and Huang, Chen and Hu, Xin and Shen, Xinke and Zhang, Dan},
  journal = {Sci. Data},
  volume  = {10},
  pages   = {740},
  year    = {2023}
}

@article{shin-j-28,
  author  = {Shin, Jaeyoung and von L{\"u}hmann, Alexander and Kim, Do-Won and Mehnert, Jan and Hwang, Han-Jeong and M{\"u}ller, Klaus-Robert},
  title   = {Simultaneous acquisition of {{EEG}} and {NIRS} during cognitive tasks for an open access dataset},
  journal = {Sci. Data},
  volume  = {5},
  pages   = {180003},
  year    = {2018},
  doi     = {10.1038/sdata.2018.3}
}

@article{hinss-m,
  title   = {Open multi-session and multi-task {{EEG}} cognitive dataset for passive brain-computer interface applications},
  author  = {Hinss, Marcel F and Jahanpour, Emilie S and Somon, Bertille and Pluchon, Lou and Dehais, Fr{\'e}d{\'e}ric and Roy, Rapha{\"e}lle N},
  journal = {Sci. Data},
  volume  = {10},
  pages   = {85},
  year    = {2023}
}

@article{gebodh-n,
  title   = {Dataset of concurrent {{EEG}}, {ECG}, and behavior with multiple doses of transcranial electrical stimulation},
  author  = {Gebodh, Nigel and Esmaeilpour, Zeinab and Datta, Abhishek and Bikson, Marom},
  journal = {Sci. Data},
  volume  = {8},
  pages   = {274},
  year    = {2021}
}

@article{nieto-n,
  title   = {Thinking out loud, an open-access {{EEG}}-based {{BCI}} dataset for inner speech recognition},
  author  = {Nieto, Nicol{\'a}s and Peterson, Victoria and Rufiner, Hugo Leonardo and Kamienkowski, Juan Esteban and Spies, Ruben},
  journal = {Sci. Data},
  volume  = {9},
  pages   = {52},
  year    = {2022}
}

@article{won-k,
  title   = {{{EEG}} dataset for {RSVP} and {P300} speller brain-computer interfaces},
  author  = {Won, Kyungho and Kwon, Moonyoung and Ahn, Minkyu and Jun, Sung Chan},
  journal = {Sci. Data},
  volume  = {9},
  pages   = {388},
  year    = {2022}
}

@article{hollenstein-n,
  author  = {Hollenstein, Nora and Rotsztejn, Jonathan and Troendle, Marius and Pedroni, Andreas and Zhang, Ce and Langer, Nicolas},
  title   = {{ZuCo}, a simultaneous {{EEG}} and eye-tracking resource for natural sentence reading},
  journal = {Sci. Data},
  volume  = {5},
  pages   = {180291},
  year    = {2018},
  doi     = {10.1038/sdata.2018.291}
}

@article{liu-y,
  title   = {Lower limb motor imagery {{EEG}} dataset based on the multi-paradigm and longitudinal-training of stroke patients},
  author  = {Liu, Yuan and Gui, Zhuolan and Yan, De and Wang, Zhuang and Gao, Ruisi and Han, Ningxin and Chen, Junying and Wu, Jialing and Ming, Dong},
  journal = {Sci. Data},
  volume  = {12},
  pages   = {314},
  year    = {2025}
}

@article{wagner-j,
  title   = {High-density {{EEG}} mobile brain/body imaging data recorded during a challenging auditory gait pacing task},
  author  = {Wagner, Johanna and Martinez-Cancino, Ramon and Delorme, Arnaud and Makeig, Scott and Solis-Escalante, Teodoro and Neuper, Christa and Mueller-Putz, Gernot},
  journal = {Sci. Data},
  volume  = {6},
  pages   = {211},
  year    = {2019}
}

@article{ghosh-r,
  title   = {{SAM} 40: Dataset of 40 subject {{EEG}} recordings to monitor the induced-stress while performing {Stroop} color-word test, arithmetic task, and mirror image recognition task},
  author  = {Ghosh, Rajdeep and Deb, Nabamita and Sengupta, Kaushik and Phukan, Anurag and Choudhury, Nitin and Kashyap, Sreshtha and Phadikar, Souvik and Saha, Ramesh and Das, Pranesh and Sinha, Nidul and et al.},
  journal = {Data Brief},
  volume  = {40},
  pages   = {107772},
  year    = {2022}
}

@article{valdes-sosa,
  title   = {The {Cuban Human Brain Mapping Project}, a young and middle age population-based {{EEG}}, {MRI}, and cognition dataset},
  author  = {Valdes-Sosa, Pedro A and Galan-Garcia, Lidice and Bosch-Bayard, Jorge and Bringas-Vega, Maria L and Aubert-Vazquez, Eduardo and Rodriguez-Gil, Iris and Das, Samir and Madjar, Cecile and Virues-Alba, Trinidad and Mohades, Zia and et al.},
  journal = {Sci. Data},
  volume  = {8},
  pages   = {45},
  year    = {2021}
}

@article{mheich-a,
  title   = {{HD-EEG} for tracking sub-second brain dynamics during cognitive tasks},
  author  = {Mheich, Ahmad and Dufor, Olivier and Yassine, S and Kabbara, A and Biraben, A and Wendling, F and Hassan, M},
  journal = {Sci. Data},
  volume  = {8},
  pages   = {32},
  year    = {2021}
}

@article{liu-h,
  title   = {An {{EEG}} motor imagery dataset for brain computer interface in acute stroke patients},
  author  = {Liu, Haijie and Wei, Penghu and Wang, Haochong and Lv, Xiaodong and Duan, Wei and Li, Meijie and Zhao, Yan and Wang, Qingmei and Chen, Xinyuan and Shi, Gaige and et al.},
  journal = {Sci. Data},
  volume  = {11},
  pages   = {131},
  year    = {2024}
}

@article{choi-g,
  title   = {{{EEG}} dataset for the recognition of different emotions induced in voice-user interaction},
  author  = {Choi, Ga-Young and Shin, Jong-Gyu and Lee, Ji-Yoon and Lee, Jun-Seok and Heo, In-Seok and Yoon, Ha-Yeong and Lim, Wansu and Jeong, Jin-Woo and Kim, Sang-Ho and Hwang, Han-Jeong},
  journal = {Sci. Data},
  volume  = {11},
  pages   = {1084},
  year    = {2024}
}

@article{wei-x,
  title   = {{ANPHY-Sleep}: an open sleep database from healthy adults using high-density scalp electroencephalogram},
  author  = {Wei, Xiaoyan and Avigdor, Tamir and Ho, Alyssa and Minato, Erica and Garcia-Asensi, Alfonso and Royer, Jessica and Wang, Yingqi Laetitia and Travnicek, Vojtech and Schiller, Katharina and Bernhardt, Boris C and et al.},
  journal = {Sci. Data},
  volume  = {11},
  pages   = {896},
  year    = {2024}
}

@article{gu-m,
  title   = {An open dataset for human {{SSVEP}}s in the frequency range of 1-60 {Hz}},
  author  = {Gu, Meng and Pei, Weihua and Gao, Xiaorong and Wang, Yijun},
  journal = {Sci. Data},
  volume  = {11},
  pages   = {196},
  year    = {2024}
}

@article{mumtaz-w,
  title   = {A wavelet-based technique to predict treatment outcome for major depressive disorder},
  author  = {Mumtaz, Wajid and Xia, Likun and Mohd Yasin, Mohd Azhar and Azhar Ali, Syed Saad and Malik, Aamir Saeed},
  journal = {PLoS ONE},
  volume  = {12},
  number  = {2},
  pages   = {e0171409},
  year    = {2017}
}

@article{iwama-s,
  title   = {High-density scalp electroencephalogram dataset during sensorimotor rhythm-based brain-computer interfacing},
  author  = {Iwama, Seitaro and Morishige, Masumi and Kodama, Midori and Takahashi, Yoshikazu and Hirose, Ryotaro and Ushiba, Junichi},
  journal = {Sci. Data},
  volume  = {10},
  pages   = {385},
  year    = {2023}
}

@article{simistira-l,
  title   = {Bimodal electroencephalography-functional magnetic resonance imaging dataset for inner-speech recognition},
  author  = {Simistira Liwicki, Foteini and Gupta, Vibha and Saini, Rajkumar and De, Kanjar and Abid, Nosheen and Rakesh, Sumit and Wellington, Scott and Wilson, Holly and Liwicki, Marcus and Eriksson, Johan},
  journal = {Sci. Data},
  volume  = {10},
  pages   = {378},
  year    = {2023}
}

@article{moffa-a,
  title   = {Neuromodulatory effects of theta burst stimulation to the prefrontal cortex},
  author  = {Moffa, Adriano H and Boonstra, Tjeerd W and Wang, Ashley and Martin, Donel and Loo, Colleen and Nikolin, Stevan},
  journal = {Sci. Data},
  volume  = {9},
  pages   = {717},
  year    = {2022}
}

@article{cuevas-r,
  title   = {An electroencephalography-based database for studying the effects of acoustic therapies for tinnitus treatment},
  author  = {Cuevas-Romero, Alma Rosa and Alonso-Valerdi, Luz Mar{\'\i}a and Intriago-Campos, Luis Alejandro and Ibarra-Z{\'a}rate, David Isaac},
  journal = {Sci. Data},
  volume  = {9},
  pages   = {500},
  year    = {2022}
}

@article{pei-x,
  title   = {A simultaneous electroencephalography and eye-tracking dataset in elite athletes during alertness and concentration tasks},
  author  = {Pei, Xinzhen and Xu, Guiying and Zhou, Yunhui and Tao, Luna and Cui, Xiaozhu and Wang, Zhenyu and Xu, Bingru and Wang, An-Li and Zhao, Xi and Dong, Haijun and et al.},
  journal = {Sci. Data},
  volume  = {9},
  pages   = {465},
  year    = {2022}
}

@article{lin-n,
  title   = {An {{EEG}} dataset for interictal epileptiform discharge with spatial distribution information},
  author  = {Lin, Nan and Zheng, Mengxuan and Li, Lian and Hu, Peng and Gao, Weifang and Sun, Heyang and Xu, Chang and Yuan, Gonglin and Liang, Zi and Dong, Yisu and et al.},
  journal = {Sci. Data},
  volume  = {12},
  pages   = {229},
  year    = {2025}
}

@article{pavlov-y,
  title   = {Pupillometry and electroencephalography in the digit span task},
  author  = {Pavlov, Yuri G and Kasanov, Dauren and Kosachenko, Alexandra I and Kotyusov, Alexander I and Busch, Niko A},
  journal = {Sci. Data},
  volume  = {9},
  pages   = {325},
  year    = {2022}
}

@article{liu-b,
  title   = {{eldBETA}: A large eldercare-oriented benchmark database of {{SSVEP}}-{{BCI}} for the aging population},
  author  = {Liu, Bingchuan and Wang, Yijun and Gao, Xiaorong and Chen, Xiaogang},
  journal = {Sci. Data},
  volume  = {9},
  pages   = {252},
  year    = {2022}
}

@article{cai-h,
  title   = {A multi-modal open dataset for mental-disorder analysis},
  author  = {Cai, Hanshu and Yuan, Zhenqin and Gao, Yiwen and Sun, Shuting and Li, Na and Tian, Fuze and Xiao, Han and Li, Jianxiu and Yang, Zhengwu and Li, Xiaowei and et al.},
  journal = {Sci. Data},
  volume  = {9},
  pages   = {178},
  year    = {2022}
}

@article{pascucci-d,
  title   = {Source imaging of high-density visual evoked potentials with multi-scale brain parcellations and connectomes},
  author  = {Pascucci, David and Tourbier, Sebastien and Ru{\'e}-Queralt, Joan and Carboni, Margherita and Hagmann, Patric and Plomp, Gijs},
  journal = {Sci. Data},
  volume  = {9},
  pages   = {9},
  year    = {2022}
}

@article{stieger-j,
  title   = {Continuous sensorimotor rhythm based brain computer interface learning in a large population},
  author  = {Stieger, James R and Engel, Stephen A and He, Bin},
  journal = {Sci. Data},
  volume  = {8},
  pages   = {98},
  year    = {2021}
}

@mastersthesis{lopez-s,
  author = {López de Diego, Silvia},
  title  = {{Automated Interpretation of Abnormal Adult Electroencephalograms}},
  school = {Temple University},
  type   = {M.S. thesis},
  year   = {2017}
}

@inproceedings{buckwalter-g,
  title     = {Recent advances in the {TUH} {{EEG}} corpus: improving the interrater agreement for artifacts and epileptiform events},
  author    = {Buckwalter, G and Chhin, S and Rahman, S and Obeid, I and Picone, J},
  booktitle = {Proc. IEEE Signal Process. Med. Biol. Symp.},
  address   = {Philadelphia, PA, USA},
  pages     = {1--3},
  year      = {2021}
}

@inproceedings{veloso-l,
  title     = {Big data resources for {{EEG}}s: Enabling deep learning research},
  author    = {Veloso, L and McHugh, J and von Weltin, Eva and Lopez, Sebas and Obeid, I and Picone, Joseph},
  booktitle = {Proc. IEEE Signal Process. Med. Biol. Symp.},
  address   = {Philadelphia, PA, USA},
  pages     = {1--3},
  year      = {2017}
}

@inproceedings{harati-a,
  title     = {Improved {{EEG}} event classification using differential energy},
  author    = {Harati, Amir and Golmohammadi, Meysam and Lopez, Silvia and Obeid, Iyad and Picone, Joseph},
  booktitle = {Proc. IEEE Signal Process. Med. Biol. Symp.},
  address   = {Philadelphia, PA, USA},
  pages     = {1--4},
  year      = {2015}
}

@article{shah-v,
  title   = {The temple university hospital seizure detection corpus},
  author  = {Shah, Vinit and Von Weltin, Eva and Lopez, Silvia and McHugh, James Riley and Veloso, Lillian and Golmohammadi, Meysam and Obeid, Iyad and Picone, Joseph},
  journal = {Front. Neuroinform.},
  volume  = {12},
  pages   = {83},
  year    = {2018}
}

@inproceedings{von-w,
  title     = {Electroencephalographic slowing: A primary source of error in automatic seizure detection},
  author    = {von Weltin, Eva and Ahsan, Tameem and Shah, Vinit and Jamshed, Dawer and Golmohammadi, Meysam and Obeid, Iyad and Picone, Joseph},
  booktitle = {Proc. IEEE Signal Process. Med. Biol. Symp.},
  address = {Philadelphia, PA, USA},
  pages     = {1--5},
  year      = {2017}
}

@article{grootswagers-t-61,
  title   = {Human infant {{EEG}} recordings for 200 object images presented in rapid visual streams},
  author  = {Grootswagers, Tijl and Quek, Genevieve L and Zeng, Zhen and Varlet, Manuel},
  journal = {Sci. Data},
  volume  = {12},
  pages   = {407},
  year    = {2025}
}

@article{yi-w,
  author  = {Yi, Weibo and Chen, Jiaming and Wang, Dan and Hu, Xinkang and Xu, Meng and Li, Fangda and Wu, Shuhan and Qian, Jin},
  title   = {A multi-modal dataset of electroencephalography and functional near-infrared spectroscopy recordings for motor imagery of multi-types of joints from unilateral upper limb},
  journal = {Sci. Data},
  volume  = {12},
  pages   = {953},
  year    = {2025},
  doi     = {10.1038/s41597-025-05286-0}
}

@article{rybavr-m,
  title   = {Simultaneous {{EEG}} and {fNIRS} recordings for semantic decoding of imagined animals and tools},
  author  = {Ryb{\'a}{\v{r}}, Milan and Poli, Riccardo and Daly, Ian},
  journal = {Sci. Data},
  volume  = {12},
  pages   = {613},
  year    = {2025}
}

@article{he-b,
  title   = {A simultaneous {{EEG}} and eye-tracking dataset for remote sensing object detection},
  author  = {He, Bing and Zhang, Hongqiang and Qin, Tong and Shi, Bowen and Wang, Qiao and Dong, Weihua},
  journal = {Sci. Data},
  volume  = {12},
  pages   = {651},
  year    = {2025}
}

@article{xue-s,
  title   = {A multi-subject and multi-session {{EEG}} dataset for modelling human visual object recognition},
  author  = {Xue, Shuning and Jin, Bu and Jiang, Jie and Guo, Longteng and Zhou, Jin and Wang, Changyong and Liu, Jing},
  journal = {Sci. Data},
  volume  = {12},
  pages   = {663},
  year    = {2025}
}

@article{bai-y,
  title   = {{TMNRED}, a {Chinese} language {{EEG}} dataset for fuzzy semantic target identification in natural reading environments},
  author  = {Bai, Yanru and Tang, Qi and Zhao, Ran and Liu, Hongxing and Zhang, Shuming and Guo, Mingkun and Guo, Minghan and Wang, Junjie and Wang, Changjian and Xing, Mu and et al.},
  journal = {Sci. Data},
  volume  = {12},
  pages   = {701},
  year    = {2025}
}

@article{moreira-j,
  author  = {Moreira, Jo{\~a}o Pedro Carvalho and Carvalho, Vin{\'\i}cius Rezende and Mendes, Eduardo Mazoni Andrade Mar{\c{c}}al and Fallah, Aria and Sejnowski, Terrence J. and Lainscsek, Claudia and Comstock, Lindy},
  title   = {An open-access {{EEG}} dataset for speech decoding: Exploring the role of articulation and coarticulation},
  journal = {Sci. Data},
  volume  = {12},
  pages   = {1017},
  year    = {2025},
  doi     = {10.1038/s41597-025-05187-2}
}

@article{wang-q,
  author  = {Wang, Qixuan and Zhou, Qian and Ma, Zhengwu and Wang, Nan and Zhang, Tianyu and Fu, Yaoyao and Li, Jixing},
  title   = {{Le Petit Prince (LPP)} multi-talker: Naturalistic {7 T fMRI} and {{EEG}} dataset},
  journal = {Sci. Data},
  volume  = {12},
  pages   = {829},
  year    = {2025},
  doi     = {10.1038/s41597-025-05158-7}
}

@article{zhang-g,
  author  = {Zhang, Guohao and Zhou, Ming and Zhen, Shuyi and Tang, Shaohua and Li, Zheng and Zhen, Zonglei},
  title   = {A large-scale {MEG} and {{EEG}} dataset for object recognition in naturalistic scenes},
  journal = {Sci. Data},
  volume  = {12},
  pages   = {857},
  year    = {2025},
  doi     = {10.1038/s41597-025-05174-7}
}

@article{tao-x,
  title   = {A multimodal physiological dataset for driving behaviour analysis},
  author  = {Tao, Xiaoming and Gao, Dingcheng and Zhang, Wenqi and Liu, Tianqi and Du, Bing and Zhang, Shanghang and Qin, Yanjun},
  journal = {Sci. Data},
  volume  = {11},
  pages   = {378},
  year    = {2024}
}

@misc{Lopez-Larraz,
  author    = {López-Larraz, Eduardo and Sierra-Torralba, María and Clemente, Sergio and Fierro, Galit and Oriol, David and Minguez, Javier and Montesano, Luis and Klinzing, Jens G.},
  title     = {{Bitbrain Open Access Sleep Dataset}},
  year      = {2025},
  howpublished = {OpenNeuro},
  note      = {Version 1.1.1, doi: 10.18112/openneuro.ds005555.v1.1.1}
}

@article{ito2025,
  author  = {Ito, Aozora and Tanaka, Toshihisa},
  journal = {IEEE Access},
  title   = {{SleepSatelightFTC}: A lightweight and interpretable deep learning model for single-channel {{EEG}}-based sleep stage classification},
  year    = {2025},
  volume  = {13},
  pages   = {46263--46272},
  doi     = {10.1109/ACCESS.2025.3549436}
}

@article{cognitive-1,
  title   = {Characteristics of physical environments that enhance learning: A systematic review of {{EEG}}-based empirical studies},
  author  = {Hong, Yi-Kyung and Cho, Ji Young},
  journal = {J. Environ. Psychol.},
  volume  = {102},
  pages   = {102525},
  year    = {2025}
}

@article{cognitive-2,
  author  = {Zhou, Yueying and Jiang, Junji and Wang, Lijun and Liang, Shanshan and Liu, Hao},
  journal = {IEEE Access},
  title   = {Enhanced cognitive load detection in air traffic control operators using {{EEG}} and a hybrid deep learning approach},
  year    = {2025},
  volume  = {13},
  pages   = {12127--12137},
  doi     = {10.1109/ACCESS.2025.3530091}
}

@article{liu2026eeg,
  title     ={{EEG} Foundation Models: Progresses, Benchmarking, and Open Problems},
  author    ={Liu, Dingkun and Chen, Yuheng and Chen, Zhu and Cui, Zhenyao and Wen, Yaozhi and An, Jiayu and Luo, Jingwei and Wu, Dongrui},
  journal   ={arXiv:2601.17883},
  year      ={2026}
}

@inproceedings{yang2026eeg,
  title     ={Are {EEG} foundation models worth it? {Comparative} evaluation with traditional decoders in diverse {BCI} tasks},
  author    ={Yang, Liuyin and Sun, Qiang and Li, Ang and Van Hulle, Marc M},
  booktitle = {Proc. Int. Conf. Learn. Represent.},
  address   = {Rio de Janeiro, Brazil},
  year      ={2026}
}

@inproceedings{REVE,
  author    = {El Ouahidi, Yassine and Lys, Jonathan and Thölke, Philipp and
               Farrugia, Nicolas and Pasdeloup, Bastien and Gripon, Vincent and
               Jerbi, Karim and Lioi, Giulia},
  title     = {{REVE}: A Foundation Model for {EEG}: Adapting to Any Setup
               with Large-Scale Pretraining on 25,000 Subjects},
  booktitle = {Adv. Neural Inf. Process. Syst.},
  volume    = {38},
  pages     = {25583--25619},
  address   = {San Diego, CA, USA},
  year      = {2025},
  doi       = {10.52202/085713-0760}
}

@inproceedings{DIVER0,
  author    = {Han, Danny Dongyeop and Lee, Ahhyun Lucy and Lee, Taeyang and
               Gwon, Yonghyeon and Lee, Sebin and Lee, Seongjin and
               Park, David Keetae and Yoo, Shinjae and Cha, Jiook and
               Chung, Chun Kee},
  title     = {{DIVER-0}: A Fully Channel Equivariant {EEG} Foundation Model},
  booktitle = {Proc. ICML Workshop Generative AI Biol. (GenBio)},
  address   = {Vancouver, BC, Canada},
  year      = {2025},
  note      = {arXiv:2507.14141},
  doi       = {10.48550/arXiv.2507.14141}
}

\end{document}